\documentclass[acmtog, authorversion]{acmart}

\AtBeginDocument{%
  \providecommand\BibTeX{{%
    \normalfont B\kern-0.5em{\scshape i\kern-0.25em b}\kern-0.8em\TeX}}}

\setcopyright{rightsretained}
\acmJournal{TOG}
\acmYear{2024} \acmVolume{43} \acmNumber{4} \acmArticle{93} \acmMonth{7}\acmDOI{10.1145/3658234}

\usepackage{pifont}

\usepackage{multirow} 
\usepackage{url} 
\usepackage{subcaption}
\usepackage{xcolor,colortbl}
\usepackage{epigraph}
\usepackage{enumitem}
\usepackage{algorithm}
\usepackage{algpseudocode}

\newcommand{\ue}{\mathcal{E}_\text{univ}}
\newcommand{\ra}[1]{\renewcommand{\arraystretch}{#1}} 

\definecolor{darkorange}{rgb}{1.0, 0.55, 0.0}
\definecolor{burntorange}{rgb}{0.8, 0.33, 0.0}
\begin{document}

\title{Universal Facial Encoding of Codec Avatars from VR Headsets}

\author{Shaojie Bai}
\email{shaojieb@meta.com}
\author{Te-Li Wang}
\email{teli@meta.com}
\author{Chenghui Li}
\email{leochli@meta.com}
\author{Akshay Venkatesh}
\email{akshayv347@meta.com}
\author{Tomas Simon}
\email{tsimon@meta.com}
\author{Chen Cao}
\email{chencao@meta.com}
\author{Gabriel Schwartz}
\email{gbscwartz@meta.com}
\author{Ryan Wrench}
\email{rpw@meta.com}
\author{Jason Saragih}
\email{jsaragih@meta.com}
\author{Yaser Sheikh}
\email{yasers@meta.com}
\author{Shih-En Wei}
\email{swei@meta.com}
\affiliation{%
  \institution{\\ Codec Avatars Lab, Meta}
  \country{USA}
}
\settopmatter{printacmref=false}

\renewcommand{\shortauthors}{Bai et al.}

\begin{abstract}
Faithful real-time facial animation is essential for avatar-mediated telepresence in Virtual Reality (VR). To emulate authentic communication, avatar animation needs to be efficient and accurate: able to capture both extreme and subtle expressions within a few milliseconds to sustain the rhythm of natural conversations. The oblique and incomplete views of the face, variability in the donning of headsets, and illumination variation due to the environment are some of the unique challenges in generalization to unseen faces. In this paper, we present a method that can animate a photorealistic avatar in realtime from head-mounted cameras (HMCs) on a consumer VR headset. We present a self-supervised learning approach, based on a cross-view reconstruction objective, that enables generalization to unseen users. We present a lightweight expression calibration mechanism that increases accuracy with minimal additional cost to run-time efficiency. We present an improved parameterization for precise ground-truth generation that provides robustness to environmental variation. The resulting system produces accurate facial animation for unseen users wearing VR headsets in realtime. We compare our approach to prior face-encoding methods demonstrating significant improvements in both quantitative metrics and qualitative results. 
\end{abstract}




\begin{teaserfigure}
  \centering
  \includegraphics[width=1.0\textwidth]{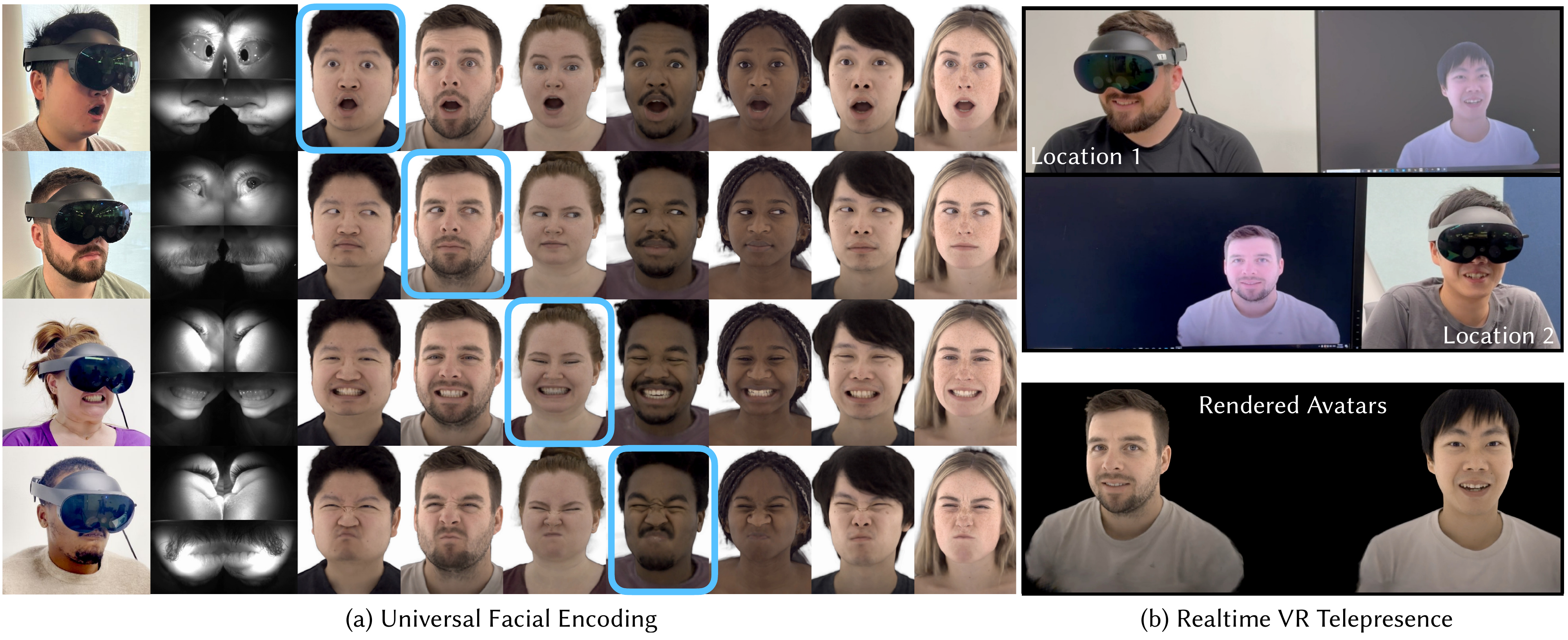}
  \caption{\textbf{Universal Facial Encoding from VR Headsets in Real-time}: We show examples of different VR users (in different rows) using the system to animate a set of universal codec avatars~\cite{cao2022authentic}. The system not only faithfully reconstructs expressions of the held-out unseen user themselves (highlighted boxes of each row), but can also transfer the expression to other identities, under challenging conditions of limited sensing.}
  \label{fig:teaser}
\end{teaserfigure}

\maketitle

\section{Introduction}
We aim to build a VR calling service that emulates in-person conversations using photorealistic avatars. An ideal VR calling system faithfully reproduces the caller's behavior by minimizing signal \emph{distortion}, emulates the immediacy of in-person conversations by minimizing \emph{latency}, and provides the affordance to have natural conversations by minimizing the \emph{interference} of the headset on caller behavior (e.g., dampening the motion of the head due to its weight). The fundamental challenge is to minimize distortion, latency, and interference in a way that is \emph{universal}: it works for the span of human diversity and across environments. Each of these objectives trade-off against the other. Reducing distortion usually involves increasing latency, since higher fidelity avatars require greater computation and necessitate increased bandwidth; slimmer VR headsets lower interference but produce greater obliquity in the viewing angles of the face and body, which leads to greater distortion. 

In this paper, we present an approach that navigates this design space to produce the first \emph{universal} face avatar encoder for photorealistic avatars from cameras on a VR headset. The avatar encoding is transmitted across the network to a receiving user's site, and input into an avatar decoder to be viewed by the receiving user in VR, in realtime. We refer to this pair of avatar encoder and decoder as a \emph{codec avatar}. For the universal avatar decoder, we build upon the \emph{universal prior model} (UPM)~\cite{cao2022authentic}, and augment it with explicit eyeball models (EEMs) and a texture branch~\cite{schwartz2020eyes}. The avatar encoder and decoder communicate via facial encodings in an expression space that is shared across identities (Fig.~\ref{fig:teaser}). We demonstrate and evaluate the codec avatars in a realtime 3D calling system on a consumer VR headset. 
\begin{figure}
  \centering
  \includegraphics[width=1.0\columnwidth]{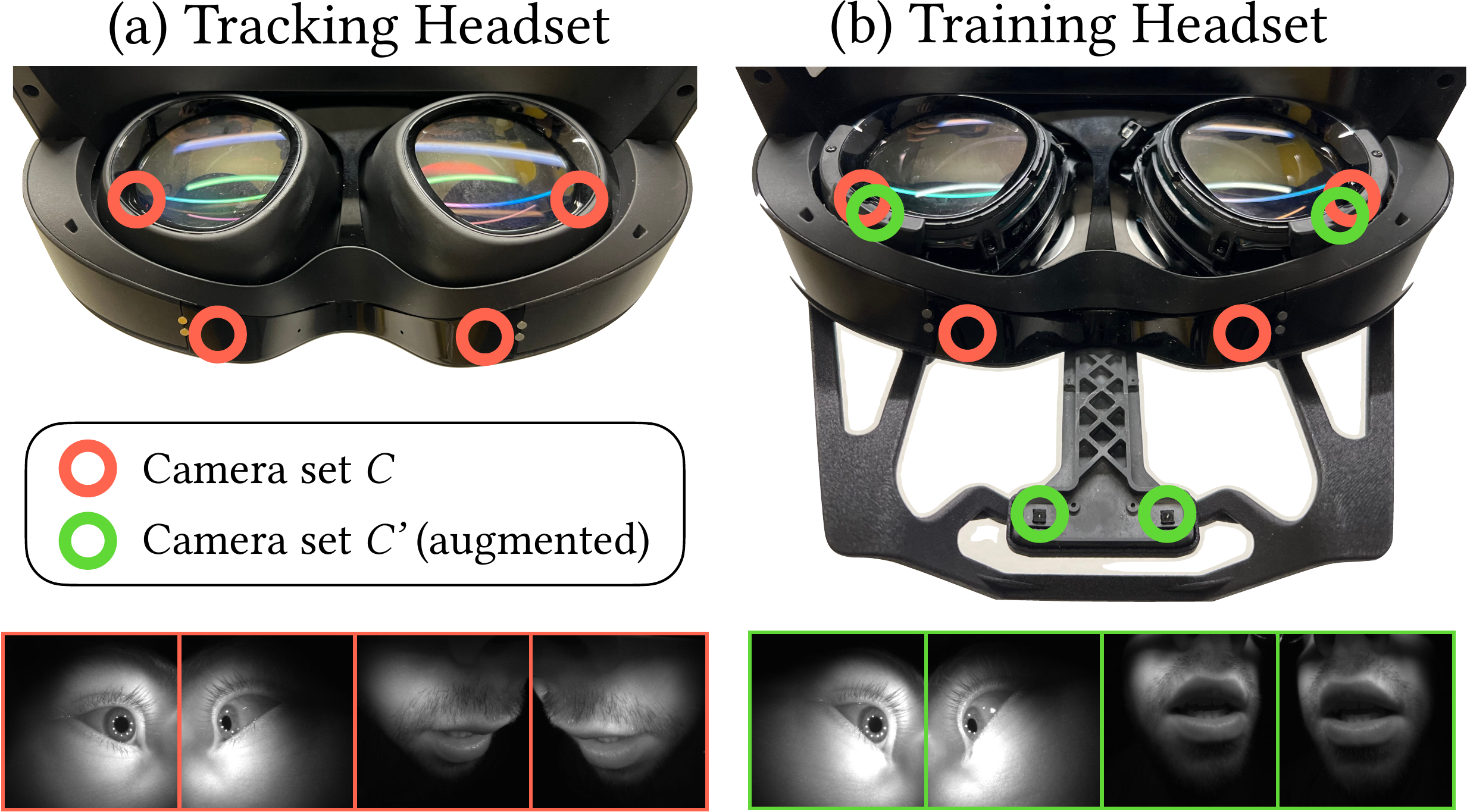}
  \caption{\textbf{Camera placements of the VR headset and example views}. (a) \textbf{Tracking headset}: an off-the-shelf VR headset that the universal encoder uses during live calls. The oblique mouth views are challenging for extracting precise facial motion; (b) \textbf{Training headset}: we augment the tracking headset with additional cameras to establish facial encoding that correspond to the training views for training encoders.}
  \label{fig:intro-headsets}
\end{figure}
The challenge in balancing the distortion and latency in facial behavior estimation emerges from the need to efficiently capture both extreme (e.g., jaw drop) and subtle expressions (e.g., a slight smirk) from an incomplete patchwork of input data. The lack of accurate facial encoding can severely disrupt the exchange of information in a conversation~\cite{ambadar2005deciphering, bould2008recognising}. Many facial encoding systems in prior literature consider non-photorealistic or stylized avatars~\cite{Hao2015,Olszewski2016,HTC_VIVE,aura23tracking}, which limit the sensitivity of users to distortion of user behavior but also limit their utility in communication~\cite{kokkinara2015animation}. The disruptive effect of latency on the efficiency of verbal communication has been established by Krauss et al.~\citet{krauss1967effects} and has been studied in the case of telepresence as well~\cite{geelhoed2009effects}.

The challenge in minimizing headset interference emerges from the need to display 3D information to the user while minimizing the weight and moment of the headset. These requirements severely constrain the available points of view from which to sense the caller's facial behavior. Most prior approaches to facial encoding from headmounted cameras have assumed clear visibility of faces~\cite{chen20223d,digidoug,chen2021high}, either making them unusable for two-way conversations or severely interfering with caller behavior. Wei et al. and Schwartz et al.~\citet{wei2019VR, schwartz2020eyes} demonstrated person-specific facial encoding on VR headsets with the notion of a \emph{training} and a \emph{tracking} headset (see Fig.~\ref{fig:intro-headsets}). The training headset is equipped with augmented cameras to enable pixel-level face registration through style transfer via the training headset. The resulting model was later distilled to work on the subset of these cameras available on the tracking headset. However, for each new person, these approaches required extended training captures and model training. Furthermore, these encoders overfit to the personalized training setup (often producing spurious correlations). Even small changes in headset donning, environmental illumination, or facial appearance significantly degraded the quality of such person-specific facial encoders. \\

\noindent \textbf{Summary of Contributions.} To achieve \emph{universality} in our encoders, we present three technical contributions. First, we present a self-supervised learning (SSL) scheme that utilizes a large collection of \emph{unpaired} HMC captures (i.e., subjects without a 3D face model), which can be collected much more rapidly. We introduce a novel-view reconstruction pretext objective that drives the learning of identity-agnostic facial expression features, leveraging the synchronization of different camera viewpoints from the HMC. This objective allows us to pre-train the facial encoder to learn robust self-supervised expression features, and then finetune the encoder using HMC data that is paired with a high fidelity 3D face model.

Second, we present an algorithm to establish ground-truth supervision for HMC inputs in the form of target encodings, across identities and environments. Establishing accurate target encoding is complicated by the occlusion of the upper face due to the VR headset. To establish this supervisory data, we generalize the correspondence algorithm presented by Schwartz et al.~\shortcite{schwartz2020eyes}, extending it to account for illumination variation introduced in difference environments. We design the style transfer function to be modulated by input lighting conditions, with mechanisms to prevent it from using its increased capacity to compensate for expression error.

Finally, given these target encodings, we train an end-to-end universal avatar encoder. At runtime, the user undergoes a lightweight calibration step, where they make a few pre-defined anchor expressions (e.g., maximal jaw drop, widened eyes). Since these anchor images are collected in the same session as the actual VR call, they have similar imaging factors (e.g., facial appearance, donning) as the incoming frames, and thus provide useful context to constrain encoding. We introduce a feature-level calibration architecture that incurs almost no additional compute cost at inference time (i.e., no latency degradation), while minimizing distortion.

To validate these innovations, we compare our approach to prior methods on over 250 identities. Quantitatively, our universal encoding approach leads to a $>20\%$ reduction in distortion. Qualitatively, we demonstrate the facial encoders in a realtime VR calling system (Fig.~\ref{fig:teaser}).

\begin{figure*}[t]
  \centering
  \includegraphics[width=1.0\textwidth]{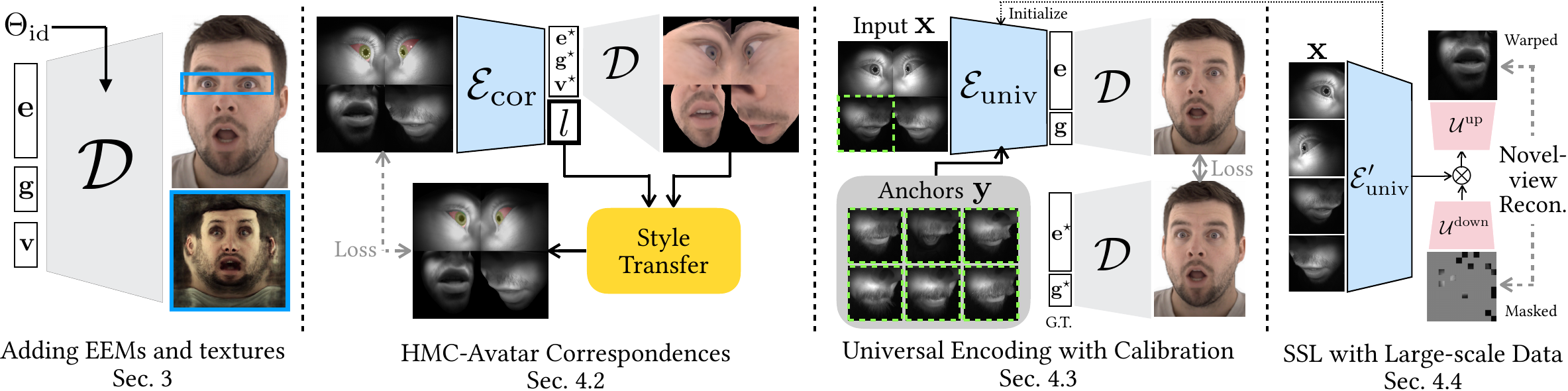}
  \caption{\textbf{System overview.} We augment universal avatar decoders to make it amenable to the style transfer method of finding HMC-avatar correspondences. We then present two methods to boost model generalization: expression calibration and self-supervised learning on unlabelled data. }
  \label{fig:intro-snapshot}
\end{figure*}
\section{Related Work}
\label{sec:relatedwork}

\subsection{Photorealistic Digital Avatars}

The generation of photorealistic digital avatars has a long history in computer graphics. One of the key drivers of advancements has been the increasing sophistication of multi-view capture systems. Increases in resolution, lighting and framerate have enabled the reconstruction of the appearance and geometry of specific individuals at high fidelity~\cite{Beeler11,Ghosh11,Klaudiny12,Bickel07,Bradley10,Furukawa09,Fyffe14,huang2006subspace,Pighin06,Zhang04}. In turn, this has allowed for the creation of accurate \emph{digital doubles} that can be driven using motion capture systems~\cite{Seymour17,Alexander10,Alexander13,Borshukov03}~. 

In recent years, techniques for digital avatar creation have gravitated towards learning-based approaches. Inspired by early work with Active Appearance Models~\cite{AAM}, techniques such as Deep Appearance Models~\cite{Lombardi2018deep} leverage the expressive power of deep neural networks to match the likeness of a person to a degree that was difficult to achieve with classical physics-inspired representations~\cite{Seymour17,cao2016real,ichim2015dynamic}. These models have been extended to incorporate explicit eyes for precise gaze control~\cite{schwartz2020eyes}, support for relighting in novel environments~\cite{bi2021avatar,yang2023towards} and adopt volumetric representations for better modeling of intricate structures like hair~\cite{rosu2022neuralstrands}. These learning-based methods also enable the creation of highly realistic representations from more constrained captures, such as from a hand-held cellphone camera~\cite{cao2022authentic,grassal2022neural}. 

For avatar-mediated interactions in VR, an important aspect of photorealistic digital avatars is their generation and rendering efficiency. Works such as~\cite{lombardi2021mixture} enable high fidelity volumetric rendering of dynamic face avatars, and the work in~\cite{remelli2022drivable} extends it to bodies. Gaussian-splatting based methods have also received a lot of attention recently owing to their ability to model intricate structures at high fidelity while affording efficient volumetric rendering~\cite{saito2023rgca}. Meanwhile, methods that specifically target on-device computation have also been explored~\cite{ma2021pixel}, enabling a self-contained compute system for avatar-mediated telepresence. Finally, concurrent to this publication, Apple Vision Pro introduces Persona (beta) application~\citep{VisionPro2024}, which enables instant avatar generation and driving.

\subsection{VR Facial Tracking and Animations}
\paragraph{Trade-off between sensing coverage and headset ergonomics.}
Tracking VR users' facial expressions is challenging due to the large occlusion from the headset. In such scenarios, the face is divided into the upper region that is enclosed by the headset, and the lower region that is still exposed to outside.
In the literature, many solutions of sensor configurations have been proposed. 
For the exposed lower face, \citet{Hao2015} and \citet{HTC_VIVE} attached the mouth sensor on a protruding mount, placing it below the mouth for a better frontal viewpoint. In~\cite{FaceVR2018}, a third-person view sensor fixed in the environment is used to avoid altering the headset. These placements either make the hardware structure fragile and imbalanced in weight, or impose constraints in a user's behavior, such as hand-face interaction and large head movements in the environment. 
For the enclosed upper face, \citet{FaceVR2018} pointed infrared (IR) cameras mounted inside the headset towards the eye areas, to avoid interfering VR usages. Despite being capable of capturing gaze direction and eye blinks, the sensor cannot capture broader upper face expressions such as movements in eyebrow, forehead, nose, and surroundings of eyes.
More recently, \citet{wei2019VR} and \citet{schwartz2020eyes} modified consumer-grade VR headsets with a mouth camera tightly attached below the headset, and highlighted the obliqueness of the camera view. Along with the two upper face cameras that cover a wide range, they built deep neural networks to directly regress expression code of photorealistic avatars. Although they obtained good tracking results, the models are highly personalized and sensitive to environmental changes.
In this work, we directly build encoding systems on an off-the-shelf VR headset, which prioritizes hardware ergonomics even more, and thus facing more challenges in sensing. Our method is not specific to any particular sensing configurations and can be directly applied to other configurations. We present methods of utilizing calibration images and large-scale data that achieve unprecedented accuracy, robustness, efficiency, and generalization to unseen users.

\paragraph{Establishing HMC-avatar correspondences}
Registering 3D face models on HMC images is a crucial step for VR face tracking.
While landmark detectors are widely used to obtain sparse geometric correspondences, as shown in~\cite{wei2019VR}, they have low expressiveness, and the reprojection error tends to fail to capture subtle movement under oblique HMC viewpoints.
Depth sensors were used in~\cite{Hao2015} and~\cite{FaceVR2018} to directly register lower mouth's geometry from an ideal camera placement. 
\citet{Olszewski2016} relied on subjects' compliance to follow known expressions and utilized audio signal to temporally align sequences to templates, resulted in low-granularity correspondences and tokenized tracking results.
With the increasing expressivity of photorealistic avatars, analysis-by-synthesis approaches \cite{Nair2008} can be used to find correspondences. However, the domain gap between HMC images and avatar renderings prohibits direct comparisons.
\citet{wei2019VR} proposed a two-stage method: they first train a CycleGAN network to convert HMC images into fake avatar renderings, before estimating expressions through differentiable renderings. This method suffered from semantic shift during the domain transfer. \citet{schwartz2020eyes} improved it by jointly learning the expression regressor and a shallow style transfer network, to avoid semantic shift during the end-to-end learning.
In this work, we find \citet{schwartz2020eyes}'s method has a severe limitation when HMC images contain large illumination variation. We significantly improved it with a design to allow style transfer network to be modulated by lighting condition in the HMC images.

\subsection{Self-Supervised Learning}

\paragraph{General self-supervised learning algorithms.} Much of the recent success in the performance of large-scale deep learning models can be attributed to the research progress on~\emph{learning from massive unlabeled data}. As opposed to supervised learning, which relies on high quality and availability of labeled data, self-supervised learning (SSL) usually defines a pretext task that allows for effective feature representation learning from the unlabeled data itself. A notable example is language modeling tasks~\citep{jozefowicz2016exploring,dauphin2017language,devlin2018bert}, where missing word/sub-word tokens are predicted from historical or surrounding natural language context. In computer vision, SSL has proven to be a strong representation learning prior in numerous contexts as well. For instance, earlier methods like~\citet{doersch2015unsupervised,noroozi2016unsupervised} have relied on relative patch positions as a supervision for local image feature learning. Another school of thought is for restoring missing information purposefully removed from the original data, such as re-colorization of images that are converted to grayscale~\citep{zhang2016colorful,larsson2016learning,vondrick2018tracking} and image inpainting~\citep{pathak2016context,liu2018image,iizuka2017globally,yu2019free}. This latter idea has recently evolved into the masked autoencoder (MAE) algorithm~\citep{he2022masked}, which aims to recover a large portion of missing patches from visible ones. Meanwhile, a completely different approach that leverages \emph{contrastive learning} has gained enormous attention. The key idea is to encourage a model to encode a pair of similar inputs (e.g., different views of the same image) into similar representations (i.e., ``positive examples'') and otherwise as different representations as possible (i.e., ``negative examples'')~\citep{jaiswal2020survey,chen2020simple,zbontar2021barlow}. Despite their strong downstream performances, these contrastive learning methods typically need heavy compute due to the requirement of large training batch sizes.

\paragraph{Self-supervision on facial representations.} Prior work has also explored numerous aspects of applying SSL on facial data. A very common scheme is self-supervised 3D face reconstruction to fit a 3D morphable model of faces (3DMM) to shape (i.e., learning 3D facial parameters from 2D single-view images)~\citep{richardson2017learning,tewari2017mofa,tewari2018self,wen2021self}. Relatedly,~\citet{he2022enhancing} proposes to use self-supervised 3D reconstruction to extract facial recognition features robust to pose and illumination.~\citet{athar2020self} uses SSL for facial expression editing, aiming to supervise motion information from image deformation. Meanwhile, prior works have explored specifically to leverage extra modalities such as audio to improve speech-related lip movement synthesis~\citep{shukla2020visually} in a self-supervised way. While these works have produced photorealistic synthesis results, they have unanimously taken advantage of some important properties of these data, such as good frontal view (i.e., most or all of face is present) and outside-in RGB input with good registration--- none of which are present in consumer VR headset (i.e., incomplete observation, inside-in monocular input, oblique camera view, extreme donning and illumination variances). To the best of the authors' knowledge, there has been no prior work in exploring a self-supervised learning approach directly from head-mounted cameras (HMC) of VR devices and demonstrating their efficacy in downstream driving of metric, photorealistic avatars with high fidelity.
\section{Overview: Codec Avatars}
\label{sec:decoder}

\begin{figure*}[htbp]
    \centering
    \begin{subfigure}[b]{0.223\textwidth}
        \centering
        \includegraphics[width=\textwidth]{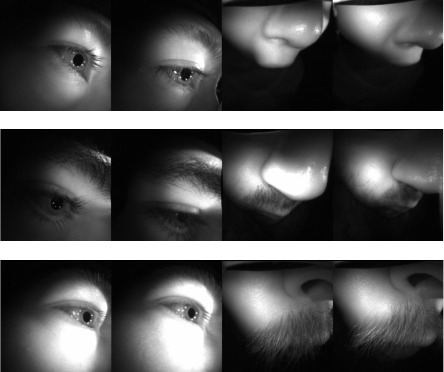}
        \caption{Incomplete observations (ours)}
        \label{subfig:hmc-occlusion}
    \end{subfigure}
    ~
    \begin{subfigure}[b]{0.223\textwidth}
        \centering
        \includegraphics[width=\textwidth]{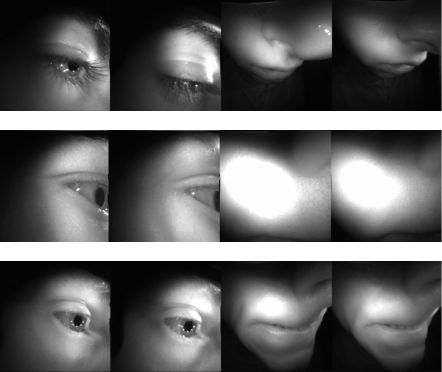}
        \caption{Donning variance (ours)}
        \label{subfig:hmc-donning}
    \end{subfigure}
    ~
    \begin{subfigure}[b]{0.223\textwidth}
        \centering
        \includegraphics[width=\textwidth]{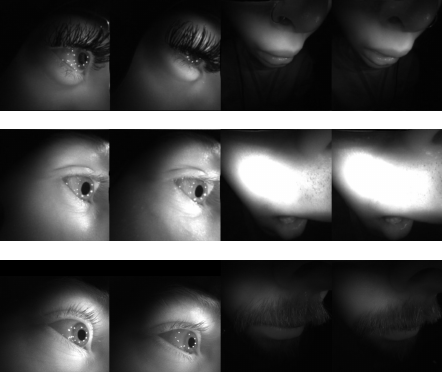}
        \caption{Illumination variance (ours)}
        \label{subfig:hmc-illumination}
    \end{subfigure}
    ~
    \begin{subfigure}[b]{0.139\textwidth}
        \centering
        \includegraphics[width=\textwidth]{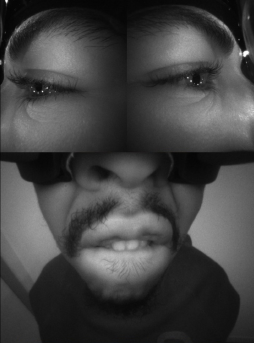}
        \caption{\citet{wei2019VR}}
        \label{subfig:hmc-alpp}
    \end{subfigure}
    ~
    \begin{subfigure}[b]{0.16\textwidth}
        \centering
        \includegraphics[width=\textwidth]{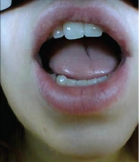}
        \caption{\citet{Olszewski2016}}
        \label{subfig:hmc-usc}
    \end{subfigure}
    \vspace{-.1in}
    \caption{\textbf{Sensing challenges} in the HMC images from the off-the-shelf VR headset we use in this work~\citep{questpro} (subfigure (a)-(c)). Compared to those used in prior works (subfigure (d) and (e)), our data suffers from more obliqueness, incompleteness, and extreme illumination.}
    \label{fig:ue-hmc-example}
\end{figure*}

A codec (en\textbf{co}der/\textbf{dec}oder) avatar is a pair of functions $(\mathcal{E}, \mathcal{D})$ designed to minimize the \emph{distortion} and the \emph{latency} in reconstructing the transmitting user's appearance and behavior at the receiving user's site. Distortion measures the error in reconstructing the appearance and behavior of the transmitting user (e.g., the mean-squared error of the texture and geometry of the transmitter's mesh). Latency measures the temporal delay in sensing, encoding, transmitting, decoding, and displaying the signal at the receiving user. In this paper, we present universal facial encoders that take, as input, image data from cameras on a VR headset and produce, as output, encodings of the state of the transmitting user, in a universal latent space that is independent of user identity.

Our universal facial encoders build upon the universal prior models (UPMs) for faces presented by~\citet{cao2022authentic}. 
Their codec avatar consists of an \emph{expression encoder} $\mathcal{E}_{\text{exp}}$ to generate expression codes $\mathbf{e}$ from view-averaged texture $\textbf{T}_{\textrm{exp}}$ and geometry images $\textbf{G}_{\textrm{exp}}$ registered from face images of a multi-camera system. 
A fully convolutional expression latent space was used, $\mathbf{e} \in \mathbb{R}^{4 \times 4 \times 16}$, to spatially localize the effects of each latent dimension and encourage semantic consistency across identities.
This is an important prerequisite for an accurate and universal encoding system. 
To preserve person-specific details, a hypernetwork \emph{identity encoder} $\mathcal{E}_{\text{id}}$ takes a neutral texture map $\textbf{T}_{\text{neu}}$ and a neutral geometry image $\textbf{G}_{\text{neu}}$ to produce multi-resolution bias maps $\Theta_{\text{id}}$.
These bias maps are used to parameterize a \emph{decoder} $\mathcal{D}$, which takes disentangled controls including viewpoint $\mathbf{v} \in \mathbb{R}^6$, expression $\mathbf{e}$, and gaze directions $\mathbf{g} = [\mathbf{g}_l, \mathbf{g}_r] \in \mathbb{R}^6$, to produce volumetric slabs $\mathbf{M}$~\cite{lombardi2021mixture} and geometry $\mathbf{G}$. A rendered image $I_{\text{vol}}$ can be then obtained through ray-marching $R_{\text{vol}}$. 
To summarize,
\begin{align}
\mathbf{e} &= \mathcal{E}_{\text{exp}}(\Delta \textbf{T}_{\text{exp}}, \Delta \textbf{G}_{\text{exp}}), \label{eq:uca2-mugsy-encoder} \\
\Theta_{\text{id}} &= \mathcal{E}_{\text{id}}(\textbf{T}_{\text{neu}}, \textbf{G}_{\text{neu}}),\\
\label{eqn:D}
\mathbf{M}, \mathbf{G} &\leftarrow \mathcal{D}(\mathbf{e}, \mathbf{v}, \mathbf{g}; \Theta_{\text{id}}), \\
I_{\text{vol}} &= R_{\text{vol}}(\mathbf{G}, \mathbf{M}, \mathbf{v}; \mathbf{K}),
\end{align}
where$\Delta \textbf{T}_{\text{exp}} = \textbf{T}_{\text{exp}}{-}\textbf{T}_{\text{neu}}$, $\Delta \textbf{G}_{\text{exp}} = \textbf{G}_{\text{exp}}{-}\textbf{G}_{\text{neu}}$ and $\mathbf{K}$ are intrinsic camera parameters. All of $\mathcal{E}_{\text{exp}}$, $\mathcal{E}_{\text{id}}$, and $\mathcal{D}$ have trainable parameters, and they were trained with variational autoencoder's objectives. We refer interested readers to~\citet{cao2022authentic} for details of how $\Theta_{\text{id}}$ parameterized $\mathcal{D}$ along with its own parameters.

\paragraph{Our augmentation.} 
For the purpose of live telepresence, we augmented this model with two important designs in a prior literature \cite{schwartz2020eyes}: (1) explicit eyeball models (EEMs): while \cite{cao2022authentic}'s model is already gaze-conditioned, the control is often not effectively reflected in the avatar's appearance, when a novel combination between $\mathbf{e}$ and $\mathbf{g}$ is encountered. 
We incorporate an eyeball geometry model into $\mathcal{D}$, and add eyeball parameters into $\Theta_{\text{id}}$. Enabling explicit control of eyeball rotations (directly from $\mathbf{g}$) leads to better eye contact in a multi person interaction. 
(2) In parallel to volumetric slabs $\mathbf{M}$, we integrate $\mathcal{D}$ with network modules to produce textures $\mathbf{T}$ for both face and eyes, where these new modules are similarly parameterized by the expanded $\Theta_{\text{id}}$. 
The motivation comes from the fact that 2D convolutions on textures are the core operations in the style transfer for establishing correspondence with HMC images~\cite{schwartz2020eyes}.
Overall, the augmented avatar decoding can be described as:
\begin{align}
    \mathbf{M}, \mathbf{G}, \mathbf{T} &\leftarrow \mathcal{D}(\mathbf{e}, \mathbf{v}, \mathbf{g}; \Theta_{\text{id}}), \\
    I_{\text{mesh}} &= R_{\text{mesh}}(\mathbf{G}, \mathbf{T}, \mathbf{v}; \mathbf{K}),
\end{align}
where $R_{\text{mesh}}$ is a standard mesh renderer with rasterization. Note that $I_{\text{mesh}}$ is only used for finding HMC-avatar correspondences. In real-time systems, we still use $I_{\text{vol}}$ that has better quality, and the compute of $\mathbf{T}$ can be saved.

In this paper, we focus on the analysis of generalization of the encoder, and thus leave the integration of phone-scan generated avatar as future work (see \S\ref{sec:discussion}). In the following, we assume $\Theta_{\text{id}}$ is available for each identity (both training and testing). We also provide in Appendix~\ref{appendix:table-of-symbols} a summarized table view of the symbols used in this paper (to reference different components of the encoder-decoder pipeline) for readers' reference.
\section{Universal Facial Encoding}
\label{sec:encoder}

Our goal is to build a universal facial encoder to extract accurate facial motion, generalize across demographics, and be robust against real-world input variation, under hardware's ergonomic constraints on VR headsets. 
In the following, we first emphasize in \S~\ref{subsec:sensing-challenge} the variety of challenges of the data from consumer-grade headsets by highlighting common examples (e.g., incomplete observations, illumination, variance, etc.). We then present an improved method over~\citet{schwartz2020eyes} to address these new data challenges while establishing high-quality avatar-to-HMC correspondences (which will serve as the labels) in \S~\ref{subsec:rosetta}. Importantly, in \S~\ref{subsec:multi-expression} and~\ref{subsec:ssl} we will introduce a \emph{fully end-to-end encoding architecture} that can generalize without heavy input canonicalization, enabled by our two main system designs: 1) multi-expression calibration (\S~\ref{subsec:multi-expression}) which substantially improves encoding accuracy via anchor expressions; as well as 2) large-scale self-supervised learning (\S~\ref{subsec:ssl}) from \emph{unlabeled} HMC data, which pretrains a encoding system before it is finetuned on low-resource high-quality labeled data.

\begin{figure*}
  \centering
  \includegraphics[width=1.0\textwidth]{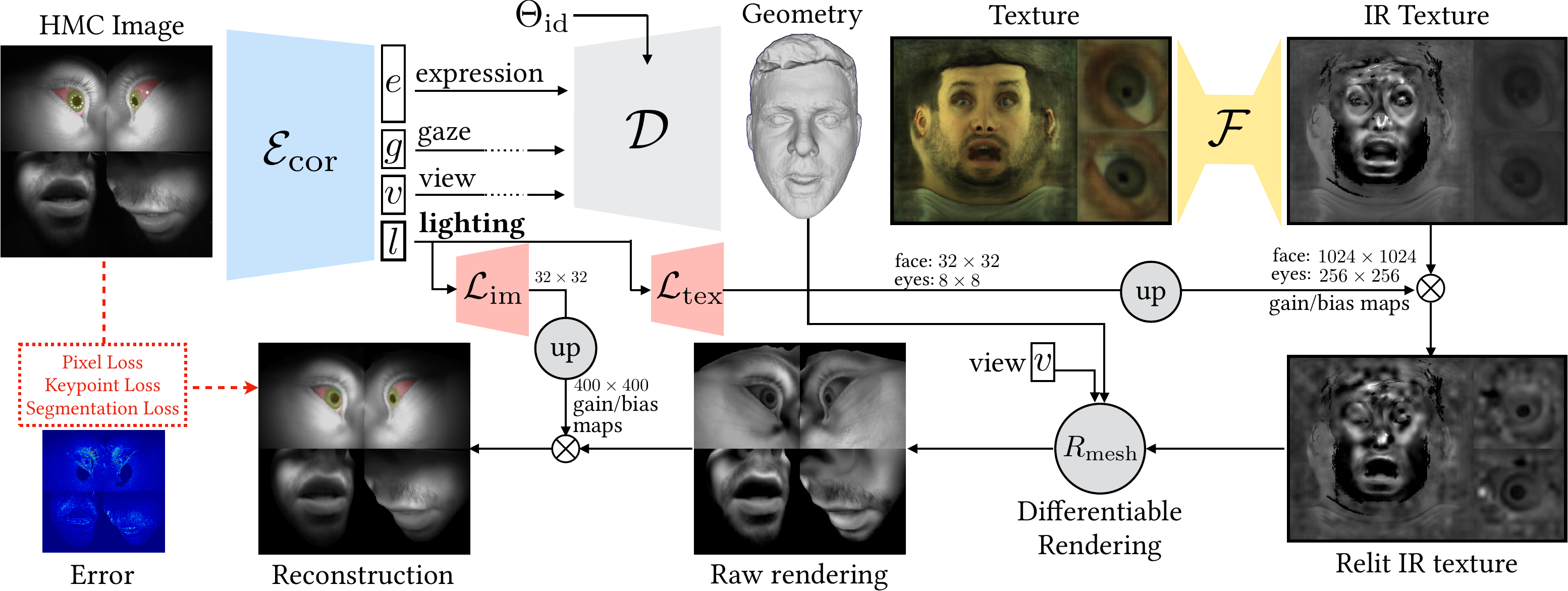}
  \caption{\textbf{Establishing HMC-avatar correspondences.} We build upon the analysis-by-synthesis method presented in prior work~\cite{schwartz2020eyes}, and address a severe limitation in handling extreme illumination changes, by regressing a ``lighting code'' to modulate style transfer in both texture and image spaces, via predicting low-resolution gain/bias maps by deconvolutional networks $\mathcal{L}_\text{im}$ and $\mathcal{L}_\text{tex}$.}
  \label{fig:ue-rosetta-architecture}
\end{figure*}
\begin{figure*}
  \centering
  \includegraphics[width=1.0\textwidth]{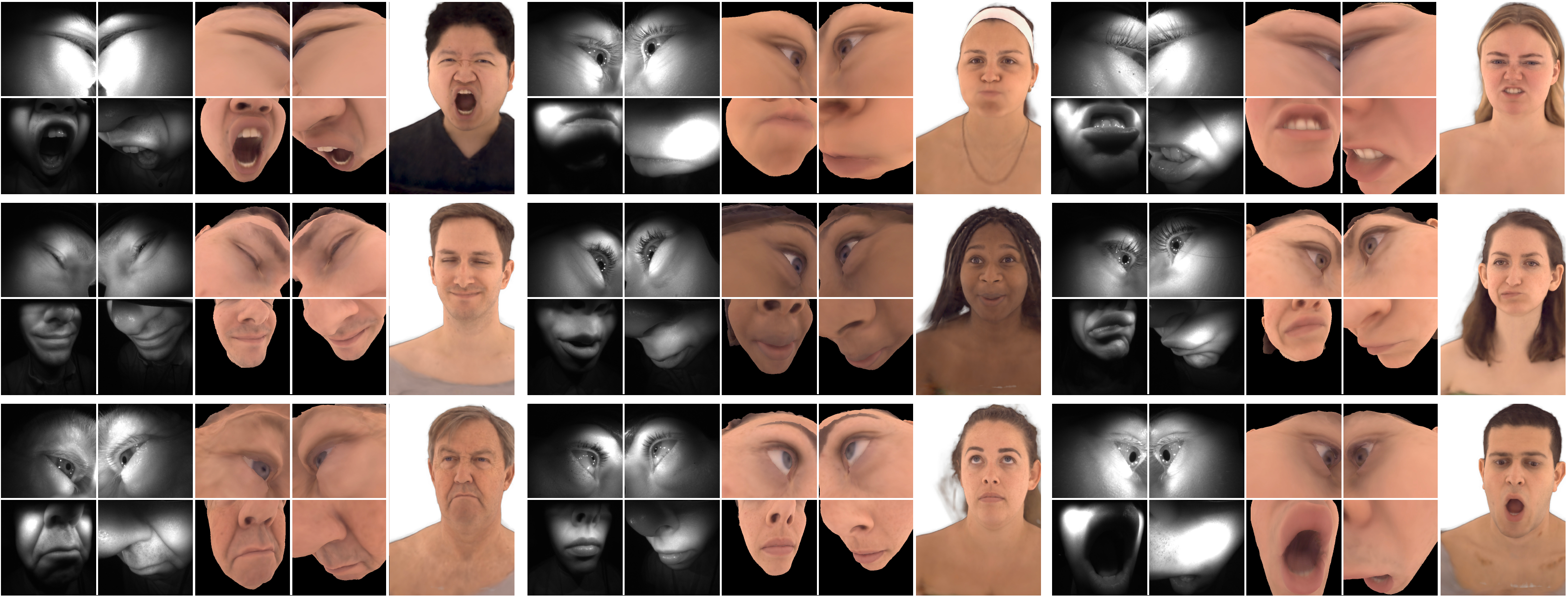}
  \caption{\textbf{Examples of established HMC-avatar correspondences.} In each grid, we show HMC images on a subset of views on the training headset (left), decoder rendering $I_{\text{mesh}}$ using found expression code $\mathbf{e}^\star$ and HMC-relative viewpoint $\mathbf{v}^\star$ with textures $\mathbf{T}$ (center), and frontal rendering $I_\text{vol}$ with volumetric slabs $\mathbf{M}$ (right).}
  \label{fig:ue-rosetta-example}
\end{figure*}

\subsection{Sensing Challenges}
\label{subsec:sensing-challenge}
In this paper, we use an off-the-shelf VR headset~\citep{questpro}. The headset has 4 inward-facing cameras, one observing each of the eyes and two observing the mouth (see Fig~\ref{fig:ue-hmc-example}(b)).
For communication, VR headset design presents a fundamental trade-off: sensing coverage against headset comfort. Better camera placements for frontal views lead to unrealistic and cumbersome VR user experience. On the other hand, making the headset compact introduces a number of sensing challenges:
\begin{enumerate}
    \item \emph{Incomplete observation.} Only part of the mouth is observed by each lower-face camera. As shown in Fig.~\ref{fig:ue-hmc-example}(a), the degree of occlusion depends on the facial features of the user. There is frequent self-occlusion due to the view angle (e.g., the upper lip or the nose blocking view of the chin) or facial hair (e.g., the mustache blocking upper lip, or even both lips).
    \item \emph{Obliqueness.} The obliquity of the camera direction with respect to the face further presents challenges as the line of sight coincides with the up/down motion of the jaw. Therefore, significant jaw motion often results in minor observed motion when projected into the images. 
    \item \emph{Donning variance.} Each time the user dons the headset, the placement of the headset with respect to the user's head is marginally different. The headset relative position and orientation can also vary within a session, as the user changes expression or comes into contact with the headset.
    The proximity of the cameras to the face has the effect that small changes in donning positions can lead to large variations in facial feature visibility (see Fig.~\ref{fig:ue-hmc-example}(b)).
    \item \emph{Illumination variance.} Due to the hardware constraints, the infrared illuminations on the headset have a rapid falloff with distance (see Fig.~\ref{fig:ue-hmc-example}(c)). With the limited dynamic range of the cameras, the images often have saturation in both dark and bright regions, correlated with donning.
\end{enumerate}
Some of these issues (e.g., incomplete observation) have been studied by~\citet{wei2019VR} to motivate the use of augmented cameras. 
The headset we use for this work added new complexities (e.g., donning and illumination variance), compared to the prototype headset used in~\citet{wei2019VR}, while making the other axes (e.g., obliqueness) worse.
We provide more details of the data we use for model training and evaluation in \S\ref{sec:experiments}.

\subsection{Groundtruth Generation}
\label{subsec:rosetta}

At each time instance (frame), the HMC produces $|C|$ synchronized camera views, $\mathbf{x} = \{\mathbf{x}_c\}_{c\in C}$.
To build a face encoder, we require the corresponding regression targets, $[\mathbf{e}^\star, \mathbf{g}^\star]$, for every frame of the training data (see Eqn.~\ref{eqn:D}). Acquiring this groundtruth is a uniquely challenging face registration problem with VR headsets, where a significant portion of the face is occluded and cannot be simultaneously observed by other cameras.
We build upon prior work~\cite{wei2019VR, schwartz2020eyes} that use analysis-by-synthesis on images from a ``training headset'' with augmented cameras that better observes the frontal face. 
The idea is that when the decoder is conditioned on groundtruth controls, the outputs of $\mathcal{D}(\mathbf{e}^\star, \mathbf{v}^\star, \mathbf{g}^\star; \Theta_\text{id})$ should generate renderings that reconstruct $\mathbf{x}_c$ after a style transfer $\mathcal{F}$ compensates for the domain difference between HMC images and decoder renderings (e.g. IR vs RGB).
The method uses a convolutional regressor $\mathcal{E}_\text{cor}$ to estimate $\mathbf{e}$, $\mathbf{v}$, and $\mathbf{g}$ directly from all HMC views, including those from the augmented cameras.

Although the augmented cameras are more amicable to visibility, all views suffer from larger variation in lighting compared with prior work (see Fig.~\ref{fig:ue-hmc-example}).
This increased challenge exposes a limitation of the method of Schwartz et al.~\shortcite{schwartz2020eyes}: the input to the style transfer module $\mathcal{F}$ is only the decoder texture, which does not carry any information about the lighting in the HMC images. This can lead to high reconstruction error and inaccurate estimates of the face state. 

To address this limitation, our $\mathcal{E}_\text{cor}$ additionally outputs a low dimensional ``lighting code'' $\mathbf{l}_c$ for each camera $c$ that modulates style transfer according to the environmental lighting condition observed the input HMC images. Specifically, we use additional deconvolutional networks $\mathcal{L}_\text{im}$ and $\mathcal{L}_\text{tex}$ to transform $\mathbf{l}_c$ into per-view gain and bias maps for the texture and image, respectively.
Fig.~\ref{fig:ue-rosetta-architecture} illustrates the overall architecture.
The two modules together better account for effects that depend on the facial expression and geometry (via the texture) as well as global effects (via the image) compared to either one alone. The generated maps are low resolution (e.g., $32 \times 32$) and are upsampled to the target texture and image sizes. This prevents the system from ``cheating'' by using the lighting branch to compensate for errors in expression estimates from $\mathcal{E}_\text{cor}$. 
Accounting for environment lighting in this way enables the accurate estimation of expressions (see Fig.~\ref{fig:ue-rosetta-example}).
We use similar reconstruction, keypoint and eye segmentation losses as Schwartz 
 et al.~\shortcite{schwartz2020eyes}, to train all modules ($\mathcal{E}_\text{cor}$, $\mathcal{F}$, $\mathcal{L}_\text{im}$, and $\mathcal{L}_\text{tex}$) jointly, for each identity independently. After convergence, we simply run inference of $\mathcal{E}_\text{cor}$ to obtain corresponding $\mathbf{e}^\star$ and $\mathbf{g}^\star$.

\subsection{Calibration Conditioned Encoding}
\label{subsec:multi-expression}

Given a collection of input-groundtruth correspondences of identity $i$ and frame 
$j$, denoted $\{(\mathbf{x}, [\mathbf{e}^\star, \mathbf{g}^\star])_{ij}\}$, our goal is to train a \emph{universal} encoder that accurately encodes facial motion of anyone. This is possible thanks to the semantic consistency of the decoder's expression space (see \S\ref{sec:decoder}). Note that we now discard the augmented camera views and only encode from the tracking headset cameras. To simplify notation, in the following we explicitly denote $|C|$ to be 4, the number of cameras on the off-the-shelf headset we use in our experiments~\cite{questpro}.

\subsubsection{Architecture}

\begin{figure}
  \includegraphics[width=0.48\textwidth]{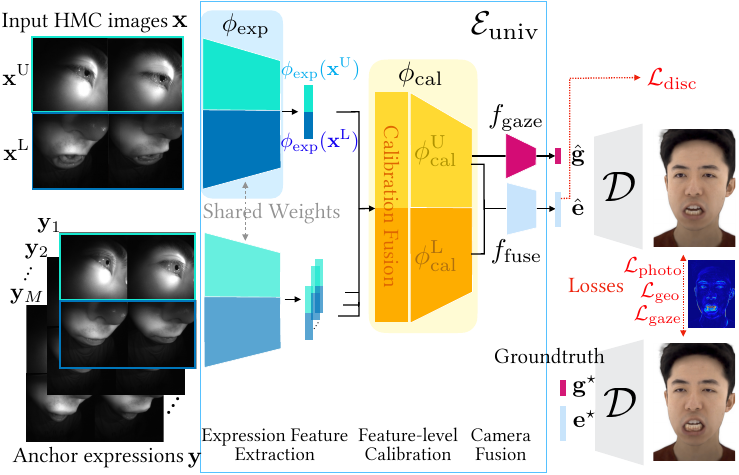}
  \caption{\textbf{Architecture and supervision of universal facial encoding with multi-expression calibration}. The architecture is composed of 3 phases: 1) expression feature extraction (\textcolor{blue}{blue region}), which computes facial features for both the current input $\mathbf{x}$ and the selected anchor expressions $\mathbf{y}$; 2) feature-level calibration (\textcolor{orange}{orange region}), which determines exact expression by calibrating current vs. anchor inputs on a feature level; and 3) camera fusion that combines multi-view camera inputs for different face regions. The architecture is trained with 4 losses: photometric, geometric, gaze and discriminative losses.
  }
  \label{fig:ue-multiexpression-architecture}
\end{figure}

As illustrated in Fig.~\ref{fig:ue-multiexpression-architecture}, we decouple the encoder architecture into three modules: (1) face-region-specific expression feature extraction $\phi_\text{exp}^{\{\text{U}, \text{L}\}}$; (2) \emph{feature-level} calibration $\phi_\text{cal}^{\{\text{U}, \text{L}\}}$; and (3) camera fusion blocks $f_\text{fuse}$ and $f_\text{gaze}$ for inferring expression and gaze, respectively, where \{U, L\} represents upper/lower face cameras.
Since the cameras are placed symmetrically on the headset (see Fig.~\ref{fig:intro-headsets}) we horizontally flip the camera images from one side of the headset to align with images from the other side (e.g., from the left eye camera's orientation to the right eye camera's). This allows the two images from the upper face cameras to share the same encoding network $\phi_\text{exp}^\text{U}$. Similarly, the two lower face cameras share the same network $\phi_\text{exp}^\text{L}$. 
This network sharing helps to avoid overfitting to the left/right correlations and increase inference efficiency via batch processing. 

Formally, we denote the current frame input as $\mathbf{x}=[\mathbf{x}^\text{U}, \mathbf{x}^\text{L}] \in \mathbb{R}^{4 \times 1 \times H \times W}$, where $\mathbf{x}^\text{U}, \mathbf{x}^\text{L} \in \mathbb{R}^{2 \times 1 \times H \times W}$ represent two upper/lower monochrome camera images flipped to the same orientation and of the size $H \times W$. 
We additionally define $M$ \emph{anchor expressions} images $\{\mathbf{y}_i\}_{i=1}^M = \{[\mathbf{y}_i^\text{U}, \mathbf{y}_i^\text{L}]\}_{i=1}^M$ where $\mathbf{y}_i \in \mathbb{R}^{4 \times 1 \times H \times W}$ represents the $i^\text{th}$ anchor expression. These anchors have predefined semantic meaning (e.g., the second one is jaw drop) and serve to calibrate the system to each person's idiosyncrasies. These same anchors were collected as part of the training set, consistently across subjects. During inference, user's go through an enrollment step where these calibration images are recorded. 
With this, we define the universal facial encoder $[\hat{\mathbf{g}}, \hat{\mathbf{e}}] = \ue(\mathbf{x})$ as follows:

\begin{align}
    \hat{\mathbf{z}}^\text{P} &= \phi_\text{cal}^\text{P}\big(\ \phi_\text{exp}^\text{P}(\mathbf{x}^\text{P}),
    \begingroup
      \color{orange}
      \underbrace{\phi_\text{exp}^\text{P}(\mathbf{y}_1^\text{P}), \ \dots, \ \phi_\text{exp}^\text{P}(\mathbf{y}_M^\text{P})}_{\text{computed only once at inference time}}
    \endgroup\big) 
    \ ; \ P \in \{U,L\}
    \label{eq:ue-multiexpression} \\
     \hat{\mathbf{e}} &= f_\text{fuse}(\hat{\mathbf{z}}^\text{U}, \hat{\mathbf{z}}^\text{L}) \\
     \hat{\mathbf{g}} &= [\hat{\mathbf{g}}_l, \hat{\mathbf{g}}_r] = f_\text{gaze}(\hat{\mathbf{z}}^\text{U}),
\end{align}
where $\hat{\mathbf{g}} \in \mathbb{R}^6$ and $\hat{\mathbf{e}}$ are the estimated gaze and expression code, respectively, and $\ue=\{\phi_\text{exp}, \phi_\text{cal}, f_\text{fuse}, f_\text{gaze}\}$ are trainable networks. 
In practice, the calibration module $\phi_\text{cal}$, which mixes current expression with anchor features, can take a few different forms. For example, it could be a self- or cross-attention Transformer that treats $\phi_\text{exp}$ as a patch embedding network~\citep{vaswani2017attention,touvron2021training,dosovitskiy2020image}, a pooling-based aggregator~\citep{qi2017pointnet} or a concatenation-based late-fusion network followed by convolutional blocks. We extensively compare these different options in \S\ref{subsec:experiments-ablations}. Empirically, we find that concatenation-based late fusion works best, and split a MobileNetV3~\citep{howard2019searching} backbone into two parts to use for $\phi_\text{exp}$ and the late-fusion portion of $\phi_\text{cal}$. We use 2-layer MLPs for both $f_\text{fuse}$ and $f_\text{gaze}$, while noting that the gaze direction prediction only uses upper-face features $\hat{\mathbf{z}}^\text{U}$. We show the encoder architecture in Fig.~\ref{fig:ue-multiexpression-architecture}.

\subsubsection{Anchor Expressions}
\label{subsubsec:anchor-expressions}

The key motivation of anchor expressions is to provide the encoder context about how a person emotes. Instead of relying solely on the \emph{absolute} configuration of facial features, the encoder can now reason based on \emph{relative} differences from known anchors. Our experiments show that this results in significantly more precise estimates of motion.

\begin{figure}
  \centering
  \includegraphics[width=0.48\textwidth]{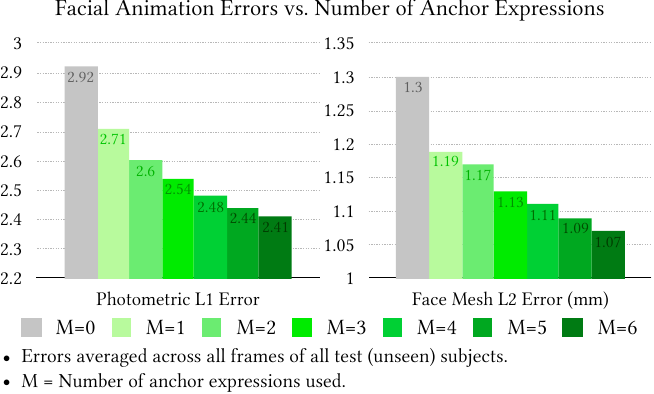}
  \caption{\textbf{Effect of multi-expression facial calibration on facial encoding model's generalization.} We use the 6 anchor expressions described in \S\ref{subsubsec:anchor-expressions} in an ablative setting. Generally, more anchor expressions lead to substantially more precise expression tracking.}
  \label{fig:ue-multiexpression-comparison}
\end{figure}

We greedily select our anchors from a set of expressions whose prompts are interpreted consistently amongst subjects, such as "neutral", "maximal jaw dropping", and "closed eyes" (See \S\ref{sec:experiments} for dataset details). The search process is based on a modified beam search~\citep{steinbiss1994improvements} algorithm. Intuitively, given an anchor set (e.g., we start from an empty set), we then identify the best subset of these aforementioned unambiguous expressions (usually the worst performing expressions by the current model) that, when added to the current anchor set, lead to the largest gain in test set performance. We iteratively repeat this process for a few iterations, keeping the best beam size of 3.  We refer interested readers to more details about our beam search algorithm in Appendix~\ref{appendix:anchor-search}. Importantly, we also perform ablative studies on the effect of using an increasing number of anchors selected this way (measured by facial animation errors), shown in Fig.~\ref{fig:ue-multiexpression-comparison}. In general, the introduction of anchor expressions leads to substantial improvement to the generalization of the encoding architecture (e.g., 18\% reduction in geometry error), while adding almost no inference time compute cost.

\begin{figure}
  \includegraphics[width=0.48\textwidth]{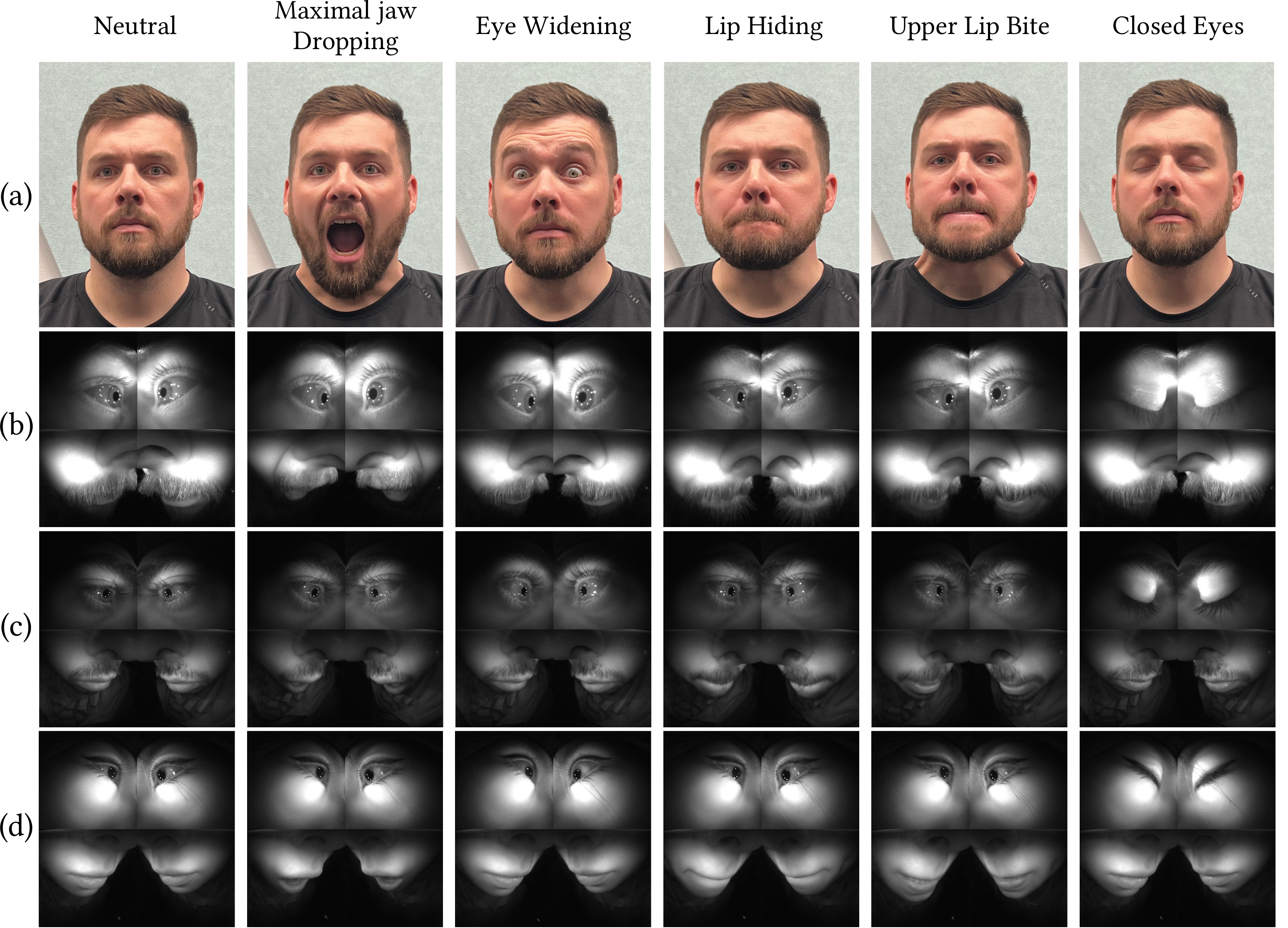}
  \caption{\textbf{Anchor expression examples.} \textbf{(a)} We show the 6 anchor expressions used for multi-expression facial calibration. \textbf{(b-d)} Examples of HMC anchor images $\mathbf{y}$ collected from 3 different subjects. }
  \label{fig:ue-multiexpression-examples}
\end{figure}

The best six anchor expressions we use in our experiments is shown in Fig.~\ref{fig:ue-multiexpression-examples}. We found that more than six anchors only brought marginal improvements empirically in our dataset.

\subsubsection{Training Loss Terms.} Our encoder training loss $\mathcal{L}$ consists of three parts. First, using the groundtruth $\mathbf{e}^\star$ and $\mathbf{g}^\star$, and a frontal viewpoint $\mathbf{v_\text{front}}$, we compute a \emph{photometric} $L_1$ loss $\mathcal{L}_\text{photo}$ directly on the volumetric rendering $I_\text{vol}$ of decoder $\mathcal{D}$.
We also penalize the error in the geometry using $\mathcal{L}_\text{geo}$ to compensate for single-view ambiguity in $\mathcal{L}_\text{photo}$:
\begin{align}
    \mathbf{M}^\star, \mathbf{G}^\star & \leftarrow \mathcal{D}(\mathbf{e}^\star, \mathbf{v}_\text{front}, \mathbf{g}^\star; \Theta_\text{id}) \\
    \hat{\mathbf{M}}, \hat{\mathbf{G}} & \leftarrow \mathcal{D}(\hat{\mathbf{e}}, \mathbf{v}_\text{front}, \hat{\mathbf{g}}; \Theta_\text{id})
\end{align}
\begin{equation}
\label{eq:photoloss}
    \mathcal{L}_\text{photo} = \left\| \mathbf{W}_1  \left( R_\text{vol}(\mathbf{G}^\star, \mathbf{M}^\star, \mathbf{v}_\text{front}) - R_\text{vol}(\hat{\mathbf{G}}, \hat{\mathbf{M}}, \mathbf{v}_\text{front}) \right) \right\|_1
\end{equation}
\begin{equation}
    \mathcal{L}_\text{geo} = \left\|\mathbf{W}_2 (\mathbf{G}^\star - \hat{\mathbf{G}})\right\|_1
\end{equation}
where $\mathbf{W}_1$ is a binary mask that only reveals a tight region covering the face, excluding regions that
are unobserved by the HMC cameras (e.g., hair, ear, and back of head).
$\mathbf{W}_2$ is an non-uniform weighting to emphasize the eye and mouth regions. Evaluating reconstruction directly on frontally rendered images more closely matches the setting of avatar-mediated communication compared to supervising with $\hat{\mathbf{e}}$ and $\hat{\mathbf{g}}$. It is made possible by virtue of the fully differentiable renderer $R_\text{vol}$ used in our decoder representation.

The second loss we employ compute the gaze error for both eyes:
\begin{align}
    \mathcal{L}_\text{gaze} = 
    \left\| \delta(\mathbf{g}^\star_{l}) - \delta(\hat{\mathbf{g}}_{l}) \right\|_2^2 + 
    \left\| \delta(\mathbf{g}^\star_{r}) - \delta(\hat{\mathbf{g}}_{r}) \right\|_2^2,
\end{align}
where $\delta(\cdot): \mathbb{R}^3 \rightarrow \mathbb{R}^2$ converts a unit-length directional vector to azimuth and altitude angles.

The third and final loss we employ 
encourages the distribution of the estimated expression codes $\hat{\mathbf{e}}$ from $\mathcal{E}_\text{univ}$ to match the distribution during decoder training, produced by $\mathcal{E}_\text{exp}$(see Eq.(\ref{eq:uca2-mugsy-encoder})).
We adversarially train a discriminator $h$ based on Wasserstein GANs with gradient penalty (WGAN-GP)~\citep{gulrajani2017improved} using:
\begin{align}
    \mathcal{L}_\text{disc} = h(\hat{\mathbf{e}}) - \underset{\mathbf{e} \sim \mathcal{E}_\text{exp}}{\mathbb{E}}[h(\mathbf{e})] + \lambda_\text{disc} \ \underset{\tilde{\mathbf{e}} \sim p_{\tilde{e}}}{\mathbb{E}}[(\|\nabla_{\tilde{\mathbf{e}}} h(\tilde{\mathbf{e}})\|_2-1)^2]
\end{align}
where $p_{\tilde{e}}$ is the distribution obtained by uniformly interpolating between $\hat{\mathbf{e}}$ and $\mathbf{e} \sim \mathcal{E}_\text{exp}$, and $\lambda_\text{disc}$ is a penalty coefficient (we use $\lambda_\text{disc}=10$). This loss prevents $\ue$ from producing codes that are far from the decoder's latent space manifold, where decoding/rendering quality tends to degrade.

In summary, our training loss for universal facial encoding is:
\begin{align}
    \mathcal{L} = \mathcal{L}_\text{photo} + \lambda_1 \mathcal{L}_\text{geo} + \lambda_2 \mathcal{L}_\text{gaze} + \lambda_3 \mathcal{L}_\text{disc}
\end{align}
where $\lambda_1, \lambda_2, \lambda_3 \in \mathbb{R}$ are weights for balancing the contributions of the different loss terms.

\subsubsection{Inference-time Complexity.} 

We collect the anchor expressions from a user only once, as a calibration process.
These images are then used repeatedly in each call to $\ue$. With reference to our encoder architecture,
we can pre-compute the feature maps $\left\{ \phi_\text{exp}(\mathbf{y}_i) \right\}_{i=1 \dots M}$ (highlighted in \textcolor{orange}{orange} in Eq.(\ref{eq:ue-multiexpression}))
only once, and re-use them in all subsequent inference calls. 
As such, in practice the use of the anchor frames 
incur \emph{negligable additional computation/speed cost} and minimal space complexity during live driving.

\subsection{Self-Supervised Facial Encoding from Large-Scale Unlabeled Data}
\label{subsec:ssl}
\begin{figure*}
  \centering
  \includegraphics[width=\textwidth]{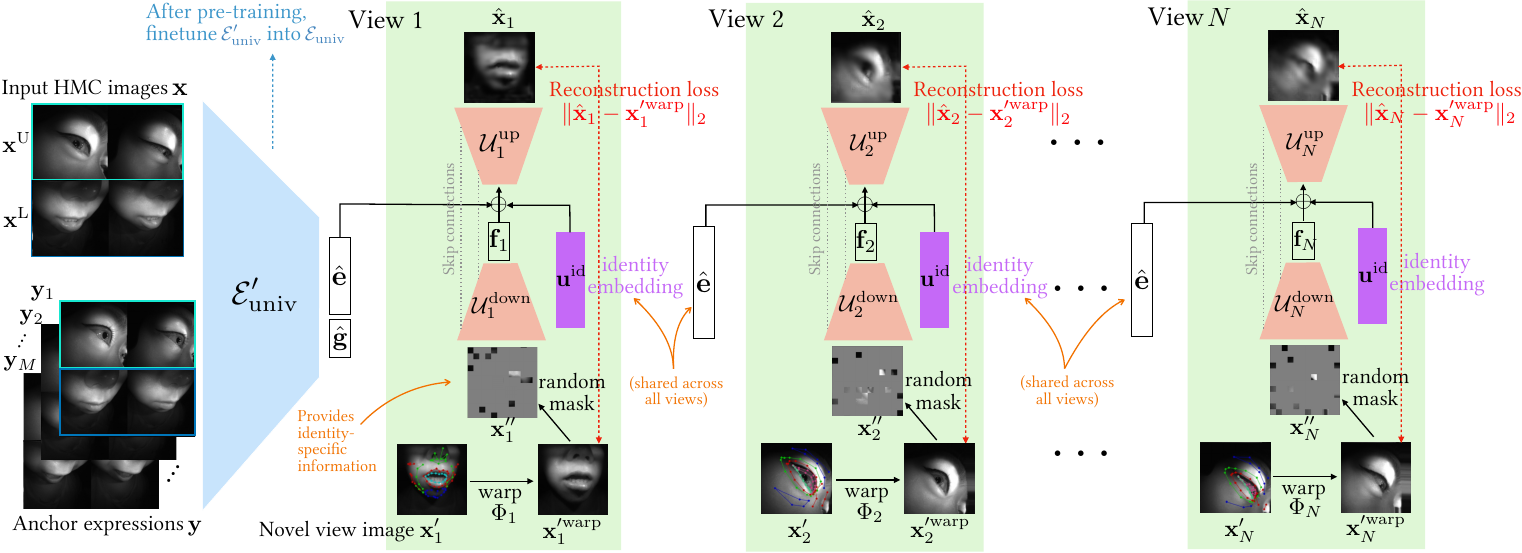}
  \caption{\textbf{Architecture of self-supervised facial encoding using unlabeled HMC data.} The leftmost module $\mathcal{E}'_\text{univ}$ is architecturally the same as $\mathcal{E}_\text{univ}$, which takes current input and anchor expressions. The encoded expression feature $\hat{\mathbf{e}}$ is then used to reconstruct the facial motion in multiple \emph{novel} views using their heavily masked inputs (e.g., 90\% of the patches were removed).}
  \label{fig:ue-ssl-architecture}
\end{figure*}

Traditionally, training facial encoding systems has relied on the ability to procure good-quality labels (e.g. blendshapes~\citep{lien2000detection,ji2022vr,aura23tracking} or learned latent space~\citep{wei2019VR,schwartz2020eyes})  with respect to a pre-existing 3D face model (e.g., 3DMM~\citep{blanz1999morphable,tran2018nonlinear} or deep appearance model~\citep{Lombardi2018deep,lombardi2021mixture}). Our method also relies on the same paradigm, as described in \S\ref{subsec:rosetta}. However, this process is both costly and highly inefficient: for each HMC captured subject, we need to collect and process a paired facial motion capture (e.g., with the multi-camera capture system~\citep{Lombardi2018deep}) to build a high-quality 3D face model. This process usually consists of multi-step optimizations; and then for each subject, we need to compute the correspondences as in \S\ref{subsec:rosetta}.

One way to resolve this challenge is to rely on unlabeled HMC data. Compared to the traditional pipeline that entails (multiple) 3rd-person-view camera(s), 3D mesh fitting, face model training, etc., HMC captures are much easier to collect (one just needs to wear a headset) and are much more lightweight to store. In this subsection, we present a novel \emph{self-supervised} learning (SSL) paradigm for facial encoding. It depends entirely on the HMC data, and defines a pretext task objective that extends the idea of masked autoencoders (MAEs)~\citep{he2022masked} by leveraging the synchronization across the multiple HMC cameras on the training headset. Overall, our SSL algorithm consists of two major components, illustrated in Fig.~\ref{fig:ue-ssl-architecture}. We discuss them separately below.

\subsubsection{HMC Expression Encoder} 
This encoder module $\mathcal{E}'_\text{univ}$ is architecturally the same as $\ue$ that we previously introduced. In particular, it takes the same inputs (i.e., current frame $\mathbf{x}$ with anchor expressions $\mathbf{y}$), and yields the same outputs $(\hat{\mathbf{e}}, \hat{\mathbf{g}})$. The goal of this module is to learn exactly the role that we expect a robust universal facial encoder would serve: extraction of expression features from HMC images that are robust to identities, incomplete observations, donning, and illumination variances.

\subsubsection{Masked autoencoders with novel-view reconstruction.} 

The idea of masked autoencoders is to learn generalizable image representations through learning how to reconstruct (intentionally created) missing patches from remaining pixels~\citep{he2022masked}. We build upon this idea and propose a self-supervision objective by leveraging the synchronization across head-mounted cameras (e.g., camera sets $C \cup C'$ for the training headset; see Fig.~\ref{fig:intro-headsets}); i.e., that different cameras contain a \emph{synchronized} multi-view of the same expression.

Specifically, given the encoded expression feature $\hat{\mathbf{e}}$ from $\mathcal{E}'_\text{univ}(\mathbf{x})$, we define a \emph{novel-view} reconstruction task that uses this expression code to reconstruct the same expression visually, but in a different HMC viewpoint. 
For example, for data captured with a training headset (with camera set $C \cup C'$), we can use the HMC images from camera set $C$, denoted $\mathbf{x}_C=\{\mathbf{x}_i\}_{i \in C}$, (which are available in off-the-shelf consumer VR headsets) to produce $\hat{\mathbf{e}} \leftarrow \mathcal{E}'_\text{univ}(\mathbf{x})$. 
Then, our goal is to use this expression code to reconstruct the augmented set camera views as accurately as possible, which we denote as $\mathbf{x}'_{C'}=\{\mathbf{x}'_i\}_{i \in C'}$. 
To achieve this, we build U-Nets $\mathcal{U}_i = \{ \mathcal{U}_i^{\text{down}}, \mathcal{U}_i^{\text{up}} \}$ with skip connections~\citep{ronneberger2015u} for each novel view $i$ (where $\mathcal{U}_i^{\text{down}}$ and $\mathcal{U}_i^{\text{up}}$ represent the downsampling and upsampling layers of a U-Net, respectively). 
Following the practice of masked autoencoders (MAE)~\cite{he2022masked}, we mask out a large portion (90\%) of the patches randomly on $\mathbf{x}'_i$, before passing it into $\mathcal{U}_i$. 
But instead of applying a vision transformer on the visible patches~\cite{he2022masked}, we concatenate an extra channel for each pixel indicating whether it's masked. 
We denote this masked and concatenated input as $\mathbf{x}''_i$, and feed it to the downsampling module $\mathcal{U}_i^{\text{down}}$ of the MAEs to obtain a $p$-channel feature map $\mathbf{f}_i \in \mathbb{R}^{p \times h \times w}$ (where $h$ and $w$ are the height and width of this feature map, respectively). Meanwhile, we learn an identity embedding $\mathbf{u}^\text{id} \in \mathbb{R}^p$ for each subject, which we update with the neural networks with gradient methods in each SSL training iteration. The upsampling module $\mathcal{U}_i^{\text{up}}$ then takes the outputs $\hat{\mathbf{e}}$ of $\mathcal{E}'_\text{univ}$, the downsampled representation $\mathbf{f}_i$, and $\mathbf{u}^\text{id}$, to reconstruct the original novel view image $\mathbf{x}'_i$. Formally, for each novel view $i$, we compute
\begin{align}
    \mathbf{f}_i & = \mathcal{U}_i^{\text{down}}(\mathbf{x}''_i), \\
    \hat{\mathbf{x}}_i & = \mathcal{U}_i^{\text{up}} \left( 
           \mathbf{f}_i, \mathcal{E}'_\text{univ}(\mathbf{x}_C), \mathbf{u}^\text{id}
        \right).
\end{align}
where we omit the notation for multi-resolution features passed through the U-Net's skip connections for simplicity~\citep{ronneberger2015u}. 

The key idea here is to learn $\mathbf{f}_i$ and $\mathbf{u}^\text{id}$ together to represent the \emph{identity}-specific information about the novel view (e.g., lighting, face shape, skin color, expression idiosyncrasies), while $\mathcal{E}'_\text{univ}$ shall encode as much identity-agnostic expression semantics as possible. We note that this structure closely resembles that of the training of the original universal facial encoding with HMC-avatar correspondences (see \S\ref{subsec:multi-expression}), where we apply $\mathcal{E}_\text{univ}$ on the HMC images $\mathbf{x}$ of the tracking headset to produce $\hat{\mathbf{e}}$, and then decode it using $\mathcal{D}$, $\Theta_\text{id}$ and viewpoint $\mathbf{v}$ into an avatar with the corresponding expression. Here, we essentially ``encode'' using $\mathcal{E}'_\text{univ}$, and then ``decode'' using $\mathcal{U}_i$, $\mathbf{u}_\text{id}$ and $\mathbf{f}_i$ to a novel view $i$'s HMC image. We repeat the decoding for all novel views $i$ that we hope to reconstruct (e.g., the 4 camera views in camera set $C'$; see Fig.~\ref{fig:intro-headsets}).

\begin{figure}
  \includegraphics[width=0.5\textwidth]{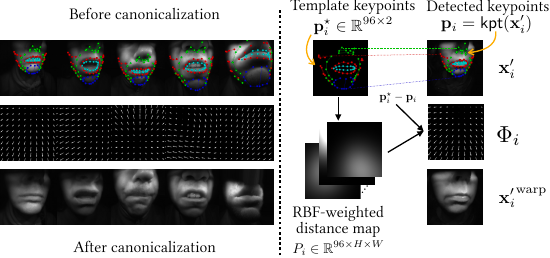}
  \caption{\textbf{HMC image canonicalization for self-supervised learning.} Before canonicalization, face shape and motion scale may be ambiguous and non-uniform (left). We use a smooth warp field to transform eyes/faces in the HMC images to similar positions, while preserving expression (right). }
  \label{fig:ue-ssl-canonicalization}
\end{figure}

However, one caveat remains with the use of augmented head-mounted cameras $C'$ for direct reconstruction. As shown in Fig.~\ref{fig:ue-ssl-canonicalization}, due to the donning flexibility of the VR headset, the absolute positions of the face in the corresponding camera frames could have a huge variance across identities. This makes the scale of facial motion ambiguous. For example, a subject whose mouth is far from the LF cameras may only have a small portion of his/her mouth in the HMC image, even if he/she performs an extreme expression like jaw drop. Then, even an \emph{utterly} wrong expression reconstruction may only incur comparatively little reconstruction error, because the mouth only occupies a small amount of pixels anyway (and vice versa). In comparison, when we train a universal facial encoder by decoding the 3D face model, we can define a fixed camera position from the center of avatar head, so that the avatars' motion scale is always consistent, instead of depending on donning/face shape. 

To solve this problem, we define a set of \emph{template keypoints} $\mathbf{p}^\star_i \in \mathbb{R}^{96 \times 2}$ for each novel view $i$, which represents a canonical scale for this camera view (i.e., 96 keypoints, with $x$- and $y$-coordinates in the image space). We then compute a distance map $P_i$ which encodes a radial basis function (RBF) weight between each pixel coordinate $(x,y) \in [W] \times [H]$ in the HMC image space and each of the 96 template keypoints $j$ of view $i$ (see Fig.~\ref{fig:ue-ssl-canonicalization}):
\begin{align}
\label{eq:ssl-canonicalization}
P_i \in \mathbb{R}^{96 \times H \times W} \ \ \text{where} \ \ {P_i}_{[j,y,x]} = \exp\left(-\frac{\|{\mathbf{p}^\star_i}_j - [x,y]\|^2}{2\sigma^2}\right)
\end{align}
where $\sigma$ controls the smoothness---i.e., a larger $\sigma$ means a pixel in the template image space could be affected more by farther keypoints. 
Meanwhile, we build an HMC image keypoint detector $\mathsf{kpt}(\cdot): \mathbb{R}^{1 \times H \times W} \rightarrow \mathbb{R}^{96 \times 2}$, and apply it on the view $i$ HMC image $\mathbf{x}'_i$: $\mathbf{p}_i = \mathsf{kpt}(\mathbf{x}'_i)$. 
We then create a warp field $\Phi_i$ by balancing $P_i$ (i.e., the relative importance of each keypoint to each pixel) and $\big[\|{\mathbf{p}_i^\star}_j - {\mathbf{p}_i}_j\|^2\big]_{j=1,\dots,96}$ (i.e., the 2D displacement of each keypoint from template) for each pixel, and use it to warp $\mathbf{x}'_i$ prior to masking. The input $\mathbf{x}''_i$ to the $\mathcal{U}_i^{\text{down}}$ is now updated to:
\begin{align}
    \mathbf{x}'^{\text{warp}}_i = \Phi_i(\mathbf{x}'_i), \\
    \mathbf{x}''_i = \mathsf{masked} \left(  \mathbf{x}'^{\text{warp}}_i \right),
\end{align}
where $\mathsf{masked}(\cdot)$ is the patch-based masking that removes a large portion of image patches and appends an additional channel of binary mask indicating whether each pixel has been masked out.
We find such an RBF-weighted warp field capable of reaching a good balance between 1) canonicalizing upper/lower-face positions in the image frame to the template; and 2) preventing large motions (e.g., jaw open max) from being overly dampened by the warp function. We demonstrate the examples of this weighted warping operation in Fig.~\ref{fig:ue-ssl-canonicalization}, where the face shapes and positions are canonicalized to similar positions after this non-linear warping.

\begin{figure}
  \centering
  \includegraphics[width=0.48\textwidth]{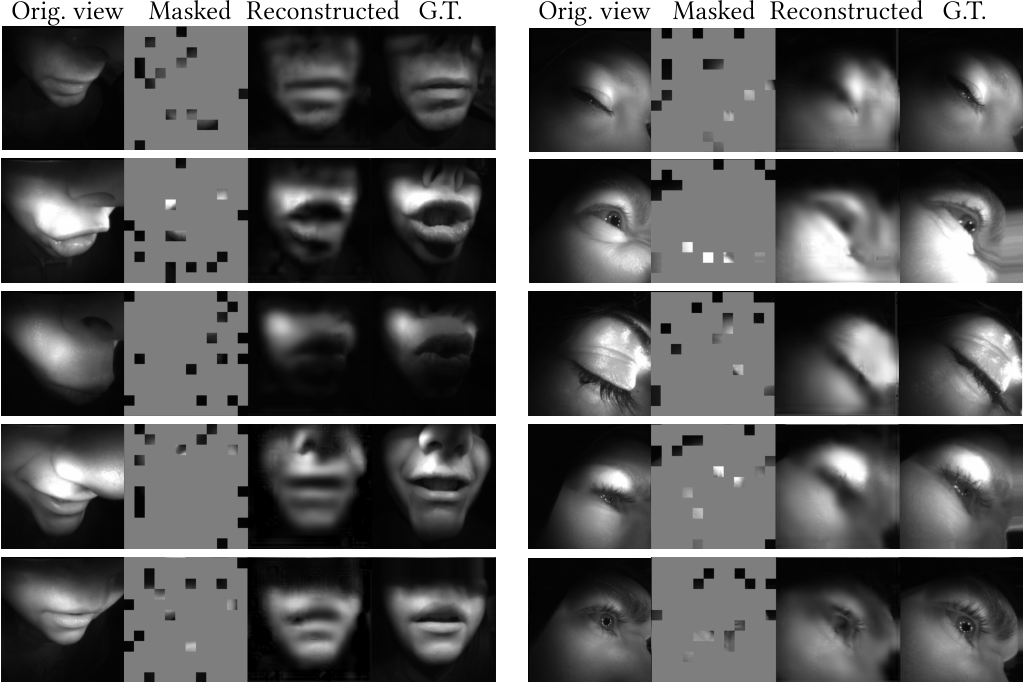}
  \caption{\textbf{Example visualizations of the self-supervised learning reconstruction.} We use images from camera set $C$ to encode expression code $\hat{\mathbf{e}}$, which the U-Nets leverage to reconstruct this expression in novel views (e.g., from camera set $C'$). As in~\citet{he2022masked}, the reconstructions are usually blurrier than the groundtruth (G.T.) due to the heavy masking. However, $\hat{\mathbf{e}}$ is able to encode expressions well for faithful reconstructions.}
  \label{fig:ue-ssl-results}
\end{figure}

\subsubsection{Training Objective.} Formally, the self-supervised learning is trained to optimize following MAE reconstruction loss~\citep{he2022masked}:
\begin{align}
    \mathcal{L}_\text{SSL} = \frac{1}{|C|}\sum_{i=1}^{|C|} \big\| 
    \hat{\mathbf{x}}_i - \mathbf{x}'^{\text{warp}}_i \big\|_2
\end{align}
We demonstrate the visualizations of SSL reconstruction results in Fig.~\ref{fig:ue-ssl-results}. As shown in the figure, with the encoded expression $\mathbf{\mathbf{e}}$, the masked autoencoders can learn to quite faithfully reconstruct the expression in a novel view setting, even though most of the patches (including those covering facial regions of interest) have been removed .

\subsubsection{Fine-tuning.} After SSL training converges, we directly use the  encoder $\mathcal{E}'_\text{univ}$ for initializing $\ue$ in the downstream task of universal facial encoding, which is a trained on a much smaller set of HMC captures, but with high-quality correspondences labels. 
Following prior work on self-supervised learning~\citep{asano2019critical,han2020self,chen2020simple,zbontar2021barlow,he2022masked}, in \S\ref{sec:experiments} we investigate both \textbf{linear probing} (i.e., hold the entire $\mathcal{E}_\text{univ}'=\ue$ fixed and train a linear layer on top of it) and \textbf{full fine-tuning} (i.e., update the parameters of the entire $\ue$) to demonstrate the generalizability of the feature extraction learned during SSL pre-training. We also compare the MAE-based SSL with a contrastive learning alternative in \S\ref{subsec:experiments-ablations}.
\section{Evaluation, Results and Qualitative Analysis}
\label{sec:experiments}

In this section, we present quantitative and qualitative studies on the performance of our approach. We start by discussing the training settings (e.g., data collection and processing) and quantitative/qualitative comparisons with competitive prior works, followed by a more in-depth study in multiple ablation settings as well as the limitations.

\subsection{Experimental Settings}
\label{subsec:experiments-settings}

\begin{table*}
\centering
\setlength{\tabcolsep}{3.5pt}
\ra{1.3} 
\caption{\textbf{Comparison of our approach with prior approaches on the held-out test subjects.} We report photometric losses (using photorealistic avatars of these unseen subjects), facial/lower-face mesh-based errors (in millimeters), and gaze direction errors (left/right eye averaged, in degrees). Inference speed measured with input image resolutions $192 \times 192$ on an NVIDIA GeForce RTX 3090 GPU with PyTorch 2.0. $\uparrow$/$\downarrow$ means higher/lower is better, respectively. \textbf{*Note}: For the mesh-tracking-based approach, we contrive a hypothetical setting to directly feed the groundtruth mesh as input (i.e., assuming perfect landmark detection and mesh tracking).}
\resizebox{\textwidth}{!}{
\begin{tabular}{l|c|ccccccc}
\toprule
Method | Metric & Input Type & Photometric $L_1$ $\downarrow$ & Mesh $L_2$ (mm) $\downarrow$ & LF Mesh $L_2$ (mm) $\downarrow$ & Mouth Shape $L_2$ (mm) $\downarrow$ & Gaze (deg) $\downarrow$ & Latency (ms) $\downarrow$ \\
\cmidrule{1-8}
Geometry-guided~\citep{song2018geometry} & raw img + heatmap & 2.978 & 1.396 & 2.356 & 3.576 & 5.01 & 10.95 \\
FLNet~\citep{gu2020flnet} & kpt displacement & 3.275 & 1.510 & 2.449 & 3.796 & 7.37 & 6.73 \\
First Order Motion Model~\citep{siarohin2019first} & kpt warp-field & 2.691 & 1.215 & 1.920 & 2.829 & \cellcolor{green!10} 4.95 & 11.73 \\
\textcolor{lightgray}{Mesh-tracking-based w/ \textbf{GT mesh}}* (see note) & \textcolor{lightgray}{tracked mesh} & \cellcolor{green!30}\textcolor{lightgray}{2.372} & \cellcolor{green!30}\textcolor{lightgray}{1.055} & \cellcolor{green!10}\textcolor{lightgray}{1.626} & \textcolor{lightgray}{2.513} & \cellcolor{green!90!black}\textcolor{gray}{0.53\textbf{*}} & \textcolor{lightgray}{N/A} \\
NeckNet~\citep{chen2021neckface} & raw img & 2.959 & 1.354 & 2.293 & 3.478 & 5.05 & \cellcolor{green!30}{4.01} \\
Meta Movement SDK~\citep{aura23tracking} & raw img & 2.942 & 1.309 & 2.202 & 3.229 & 5.15 & \cellcolor{green!90!black}\textbf{2.35} \\
\cmidrule{1-8}
\textbf{Ours} w/o SSL & raw img & \cellcolor{green!10}2.413 & \cellcolor{green!10}1.071 & \cellcolor{green!30}1.527 & \cellcolor{green!30}2.355 & \textbf{5.25} & 4.98 \\
\textbf{Ours} (SSL + linear probing) & raw img & 2.649 & 1.169 & 1.747 & \cellcolor{green!10}2.370 & 5.66  & 4.98 \\
\textbf{Ours} (SSL + full fine-tuning) & raw img & \cellcolor{green!90!black}\textbf{2.309} & \cellcolor{green!90!black}\textbf{1.023} & 
\cellcolor{green!90!black}\textbf{1.394} & 
\cellcolor{green!90!black}\textbf{2.083} & \cellcolor{green!30}\textbf{4.90} & 
\cellcolor{green!10}{4.98} \\
\bottomrule
\end{tabular}}
\label{tab:experiments-full-baseline-comparison}
\end{table*}

\paragraph{Platform.} Our experiments are conducted on NVIDIA A100 GPUs with 80GB GPU memory. The deep learning models are built with PyTorch 2.0~\citep{paszke2019pytorch} and CUDA 11.4. Specifically, the groundtruth generation pipeline (see \S\ref{subsec:rosetta}) is trained in a person-specific setting where each subject is trained individually on an A100 GPU, whereas the universal facial encoding model (as well as baseline models) is trained on all subjects with distributed data-parallel (DDP) on two A100 GPUs.

\paragraph{Data Collection.}\footnote{We release the dome-captured data and the HMC data (with labels) of these corresponding 256 subjects at \texttt{https://github.com/facebookresearch/ava-256}.} Our data consists of three sets. The \textbf{first set} is a collection of 266 subjects whose facial performance were captured in a spherical capture dome equipped with 40 color and 50 monochrome cameras that produce images of resolution $4096 \times 2668$ at 90 frames per second (we refer interested readers to~\citet{cao2022authentic} for more details). These high-resolution captures are used to train the photorealistic UPM face model augmented with explicit eyeball models (see \S\ref{sec:decoder}). The \textbf{second set} is a collection of the \emph{same} 266 subjects' head-mounted-camera (HMC) data using an augmented training headset (see Fig.~\ref{fig:intro-headsets}). For most of these subjects, their HMC data is collected on a nearby date to their capture dome data so as to ensure relatively minimal appearance change. Specifically, the HMC capture script consists of 1) 43 segments of facial expressions or their combinations (e.g., jaw open max, jaw open slightly, cheek puff, brows lowered with widened eyes, etc.; we pick the anchor expressions from these segments) with 8 seconds for each segment; 2) 21 segments of short sentence reading or natural conversation, totaling 180 seconds; 3) a free-form range-of-motion segment (for both upper and lower face) with 40 seconds. These HMC data are captured at 72 frames per second with the training headset which has 8 cameras ($C \cup C'$; see Fig.~\ref{fig:intro-headsets}) thereby leading to roughly 40K HMC frames (320K HMC monochrome images of resolution $400 \times 400$) per subject in total. The \textbf{third set} is a large-scale collection of over 17,000 subjects with the same HMC capture script (i.e., roughly 40K frames per subject), totaling over 600 million frames. However, we note that these 17K subjects do \emph{\textbf{not}} have paired high-resolution in-dome captures (i.e., no corresponding 3D face models, and thus no HMC-avatar correspondences).

\paragraph{Train \& Test data split.} For the 266 subjects with paired 3D face models, we randomly split them into 234 training subjects and 32 test subjects to evaluate the generalization of universal facial encoding on unseen facial appearances, donnings and lighting variance (i.e., neither the decoder nor the HMC images of these 32 test subjects were seen during encoder training). For the large-scale unlabeled HMC data collection, we use all of them for SSL pre-training. After pre-training, we finetune the encoder model using the aforementioned 234 training subjects.

\paragraph{Hyperparameters and Augmentation.} Our pre-training consists of 400K training steps. For training/fine-tuning universal facial encoding models, we adopt consistent hyperparameter settings in all of our experiments (unless specified otherwise). For loss penalties, we use $\lambda_1=1$ for $\mathcal{L}_\text{geo}$, $\lambda_2=20$ for $\mathcal{L}_\text{gaze}$, and $\lambda_3=0.1$ for $\mathcal{L}_\text{disc}$. We use a minibatch size of 16, weight decay $10^{-5}$ and Adam optimizer~\citep{kingma2014adam} with initial learning rate $10^{-3}$ annealed at a cosine schedule~\citep{loshchilov2016sgdr} over 150K training iterations. For augmentations, we apply random rotation (by up to $15^\circ$) and random scaling (by up to 1.1$\times$) to the HMC images at training time and use a CutMix-based~\citep{yun2019cutmix} input image mixing approach for regularization. Before feeding the $400 \times 400$ HMC images to $\mathcal{E}_\text{univ}$, we downsample them to $192 \times 192$ resolutions in all settings. We provide mor e hyperparameter details in Appendix~\ref{appendix:additional-stats}.

\paragraph{Baseline approaches.} To the best of our knowledge, we are the first to model \emph{inside-in} VR universal face encoding for photorealistic avatars. This allows us to evaluate the generalization of tracking quality in a photometric way--- in contrast to merely using a generic blendshape for every subject (e.g., Aura~\citep{aura23tracking}). We extensively compare with multiple state-of-the-art (SoTA) approaches , adopting their methodologies in our setting (with careful hyperparameter tuning to get the best performance, if needed). The latest Meta Movement SDK~\citep{aura23tracking} and NeckNet~\citep{chen2021neckface} used raw head/neck-mounted camera images as input for blendshape animations. FLNet~\citep{gu2020flnet} relied on detected keypoint/landmark locations (relative to a reference) as input representations, while ~\citet{song2018geometry} proposed to use landmark heatmap along with the image. We build an HMC keypoint detection module (which also detects pupil keypoints in order to regress gaze) to leverage these models. Alternatively, prior SOTA approaches have also used 2D keypoints to produce warping feature maps for image animation (e.g., First Order Motion models)~\citep{siarohin2019first,averbuch2017bringing}; we adapt these approaches in our context by computing the warp field between detected HMC keypoints with respect to a reference. 
Finally, a different yet highly successful approach for video reenactment is that of~\citet{thies2016face2face}, which is based on mesh tracking on RGB video. For the purpose of comparison, we approximate this method by introducing a \emph{hypothetical} setting where we \textbf{directly use groundtruth mesh} obtained from \S\ref{subsec:rosetta} (i.e., assuming \emph{perfect} keypoint detection and mesh fitting already) to regress expression/gaze codes.

\subsection{Quantitative Comparison}

The results of our approach vs. baseline encoding methods are shown in Table~\ref{tab:experiments-full-baseline-comparison}.

\paragraph{Metrics.} We exploit several metrics to evaluate the facial encoding performances, introduced below.
\begin{itemize}
    \item \emph{Photometric Image $L_1$ Error}. We directly compare and compute the pixel-level errors of the renderings of the predicted expression $[\hat{\mathbf{g}}, \hat{\mathbf{e}}]$ and the groundtruth expression $[\mathbf{g}^\star, \mathbf{e}^\star]$ on the codec avatar. To keep the metric consistent, we set viewpoint $\mathbf{v}=\mathbf{v}_\text{front}$ to be a frontal camera position for all photometric errors reported in this section.
    \item \emph{Mesh $L_2$ Error.} Average $L_2$ displacement error of the predicted vs. groundtruth face mesh in the facial region (e.g., excluding vertices on the eyes, hair, neck and shoulder). The unit of this error is millimeter (mm).
    \item \emph{Lower Face (LF) $L_2$ Error.} Similar to the mesh $L_2$ error above, except it further masks out the upper face region. The unit of this error is millimeter (mm).
    \item \emph{Mouth Shape $L_2$ Error.} Following prior work (e.g.,~\citep{bao2023learning}) we compute the horizontal and vertical $L_2$ distances between two pairs of mesh vertices (between left and right lip corners; and between centers of upper and lower lips, respectively). The two distances are compared to the groundtruth mesh and summed together. This error measures the mouth shape difference, which is especially important for speech. The unit of this error is millimeter (mm).
    \item \emph{Gaze Error.} The average of left and right eye gaze direction errors; measured in degrees.
\end{itemize}

\begin{figure*}
  \centering
  \includegraphics[width=\textwidth]{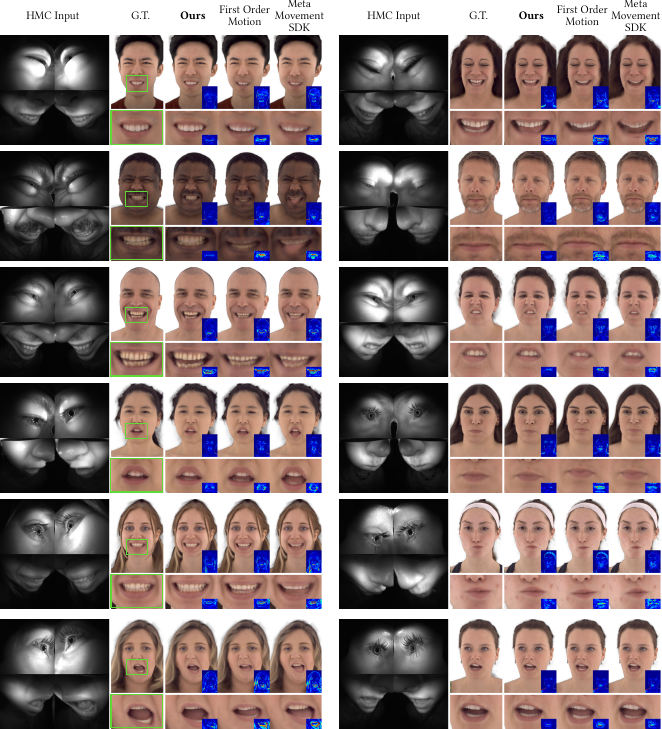}
  \caption{\textbf{Qualitative results of our method vs. prior approaches.} We show comparisons of our universal facial encoding model predictions to those of the previous methods and the corresponding groundtruth, rendered from a frontal camera view. The images are, from left to right of each column: HMC input images, Groundtruth, \textbf{ours}, keypoint warp-field-based First Order Motion methods~\citep{siarohin2019first} and Meta Movement SDK\cite{aura23tracking}. The error map of each method's prediction with respect to the groundtruth is shown in the lower right corner of the rendering.}
  \label{fig:experiments-qualitative-validation}
\end{figure*}

\begin{figure}
  \centering
  \includegraphics[width=0.47\textwidth]{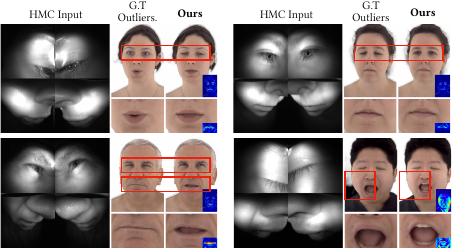}
  \caption{\textbf{Universal facial encoder can generalize well even in cases where correspondences fail.} For a small portion of the data, the style-transfer groundtruth generation pipeline (which is based on a personalized model) could fail to establish accurate correspondences. Sometimes the groundtruth code $[\mathbf{e}^\star, \mathbf{g}^\star]$ could lead to uncanny artifacts in the volumetric rendering (e.g., see bottom-right corner), as they may be out of decoder's latent expression code distribution. In comparison, our facial encoder is much more robust and produces even better results in these cases.}
  \label{fig:experiments-rosetta-outliers}
\end{figure}

\begin{figure*}
  \centering
  \includegraphics[width=\textwidth]{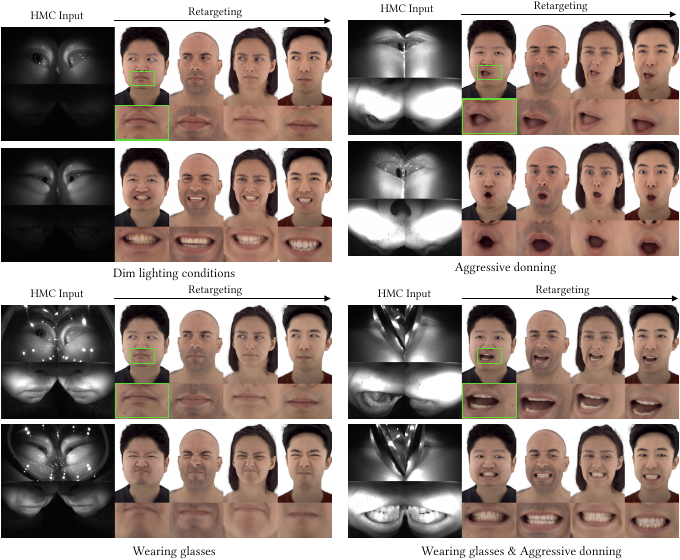}
  \caption{\textbf{Retargeting HMC expression to multiple photorealistic avatars under challenging HMC input conditions.} Even under some suboptimal or out-of-distribution scenarios (e.g., lower face camera with dim lighting conditions, or glasses), our universal facial encoding model still demonstrates relatively robust behavior that allows for expressive animations.}
  \label{fig:experiments-aggressive-comparison}
\end{figure*}

\begin{figure}
  \centering
  \includegraphics[width=0.47\textwidth]{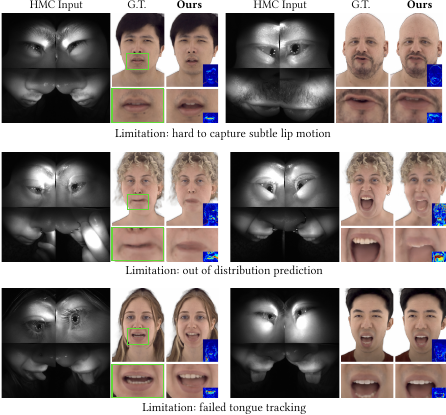}
  \caption{\textbf{Failure cases.} Our approach can still further improve on capturing subtle lip motions (which is important for speech cues) and could produce latent codes that are out-of-distribution.}
  \label{fig:experiments-limitation}
\end{figure}

\paragraph{Performance on Overall Avatar Appearance.} Photometric $L_1$ error measures the overall fidelity of avatar animation, as it most directly reflects the visual appearances of a 3D face model driven by the universal encoding system, taking into account all factors such as motion, eye gaze, skin tone, facial hair, freckles, etc. (in contrast, the geometry errors mostly measure motion and relative displacements). As shown in Table~\ref{tab:experiments-full-baseline-comparison}, our full method (SSL-pretrained multi-expression calibrated encoder) substantially improves over any prior method by over 17\%, including the existing Meta Movement SDK~\citep{aura23tracking} (2.309 vs.2.942). Surprisingly, it also surpasses the performance of the contrived setting (see ``\textcolor{lightgray}{Mesh-tracking-based w/ GT mesh}'' line in Table~\ref{tab:experiments-full-baseline-comparison}) where we directly leverage groundtruth mesh as the tracked mesh input for encoding (which has almost perfect gaze estimation).

\paragraph{Performance on Motion Geometry.} Compared to the photometric loss, the mesh-based errors can better capture absolute/relative facial motions, especially nuances (e.g., mouth shape during speech, such as lip closure). Across all three mesh-based metrics, our SSL-pretrained universal facial encoder performs best, \textbf{especially} on lower-face (LF) deformations, with an average 2.083mm error across all frames of the 32 (unseen) test subjects. This is not merely substantially lower than the existing baselines in literature, but also the facial encoding model without pre-training (i.e., the ``\textbf{Ours} w/o SSL'' row; 2.083mm vs. 2.355mm). The benefit of pre-training can be also seen in the SSL + linear probing setting: by simply fine-tuning a linear layer on top of the SSL-pretrained weights, the mouth shape error already outperforms all baselines and is on par with the supervised learning setting. We suspect this is a consequence of the novel view reconstruction in self-supervised encoder learning, where we leverage the \emph{augmented cameras} (which has a better LF view) to supervise expression encoding under incomplete observations.

\paragraph{Performance on Gaze Prediction.} The contrived setting that uses a groundtruth mesh tracking as input achieves almost perfect gaze estimation. This is because the gaze code of each eye $\mathbf{g}_\text{l}, \mathbf{g}_\text{r}$ is a 3-dimensional unit vector representing the $x-$, $y-$, $z-$axis of gaze direction (i.e., with physical meaning)~\citet{schwartz2020eyes}, which can be deduced accurately from the groundtruth explicit eye model mesh. In contrast, the expression code is a latent representation learned to encode all idiosyncrasies, semantics, etc, which is inherently ambiguous. As is shown in Table~\ref{tab:experiments-full-baseline-comparison}, the fully finetuned model obtained the second best performance, after the hypothetical set up. This again demonstrates the benefits of our self-supervision pipeline. We do note that the keypoint-based approaches tend to also have good gaze estimation, likely because of the pupil keypoints detection.

Overall, with our approach, we see consistent and significant improvements over any prior models on photorealistic facial animation. These results demonstrate a high-fidelity encoding system that is meanwhile real-time efficient. Our method can effectively extract lower face signals despite the often incomplete observations or adverse illumination, and achieves the best performance across the board.

\subsection{Qualitative Evaluation}
In addition to the several quantitative metrics, we also qualitatively visualize the improvement in the facial animation quality under various circumstances (oblique camera view, adverse lighting condition) across identities in Fig.~\ref{fig:experiments-qualitative-validation}. We specifically include the renderings of the predictions from the best two baseline approaches in Table~\ref{tab:experiments-full-baseline-comparison} (i.e., First Order Motion model~\citep{siarohin2019first}, as well as the Meta Movement SDK model~\citep{aura23tracking}). We also provide facial animation videos driven from HMC in the supplementary video.

\paragraph{Obliqueness.} Compared to prior methods, our model can regress to much more precise lower face motion even when occlusion (e.g., lips occluding chin, facial hair) is present. In comparison, the keypoint warp-field-based First Order Motion model frequently misses nuanced details (e.g., lip corner differences, or wrinkles in the glabella region), while the Meta Movement SDK model could not handle obliqueness well (as evidenced by wrong lower face geometry predicted) and is more prone to producing blurry artifacts in the volumetric facial model. Our model is also robust across settings, even in cases where the offline, personalized style-based groundtruth generation pipeline may fail (see Fig.~\ref{fig:experiments-rosetta-outliers}).

\paragraph{Donning and illumination variance.} Our model performs with consistently high fidelity across a wide spectrum of donning and illumination conditions over the baselines, as shown in Fig.~\ref{fig:experiments-qualitative-validation}. Moreover, as shown in Fig.~\ref{fig:experiments-aggressive-comparison}, the model is able to generalize to certain out-of-distribution dim lighting conditions and aggressive donning, while still maintaining expressive retargeting.

\paragraph{Glasses.} Surprisingly, our model generalizes well to certain out-of-distribution input types, such as cases where the subjects wear glasses (even though \emph{none} of the 266 subjects in our dataset with high-quality labels wear glasses during capture), see Fig.~\ref{fig:experiments-aggressive-comparison}. We hypothesize this is because 1) the large-scale pre-training enables robust eye-feature extraction; and 2) the anchor expressions provide references for relative eye motions.

\begin{figure*}
  \centering
  \includegraphics[width=\textwidth]{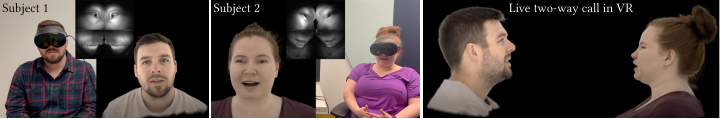}
  \caption{\textbf{2-way live call with codec avatars in VR.} We present a live demo using our approach that enable real-time communication via live driving of photorealistic avatars in VR (with an off-the-shelf VR headset~\citep{questpro}). We show a recording snippet in our supplementary video.}
  \label{fig:experiments-2way}
\end{figure*}

\begin{figure*}
  \centering
  \includegraphics[width=\textwidth]{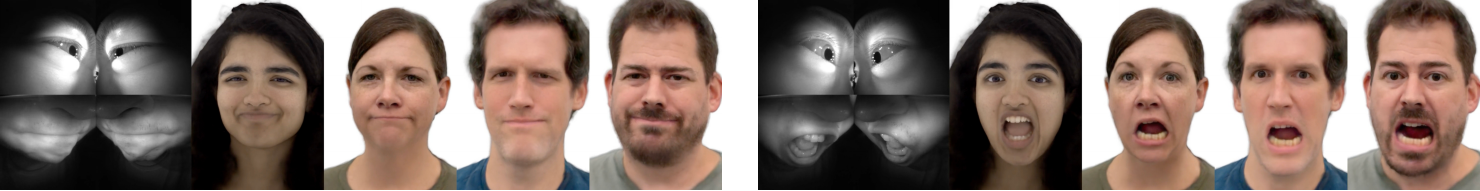}
  \caption{\textbf{Our encoder driving avatars unseen during UPM training}, from both dome (leftmost one) and phone scan captures (3 rightmost ones).}
  \label{fig:experiments-unseen}
\end{figure*}

\paragraph{Driving avatars unseen by UPM} Following~\citet{cao2022authentic} which also established the generalizability of UPM models themselves, we show in Fig.~\ref{fig:experiments-unseen} we can use the universal facial encoder to easily and accurately drive avatars that are unseen during both encoder and decoder training--- for example, avatars generated directly from a phone scan (but in the same expression latent space). 

\paragraph{Live driving.} We present a demo system that combines all of our contributions and use it to enable high-fidelity live two-way calling with (unseen) subjects driving their own photorealistic avatars (see Fig.~\ref{fig:experiments-2way} and our supplementary video).

\paragraph{Limitations.} Despite the substantial improvement of our encoding model over prior work, we still note some important failure cases/types that the system will need to further improve on (Fig.~\ref{fig:experiments-limitation}). First, subtle motion is occasionally inconsistent, such as exact lip shape. This is usually due to the inherent ambiguity caused by the incomplete observation and extreme illuminations. In practice, we found this make speech-related lip motion suboptimal at times. Second, the output latent code may be out of the decoder's pre-defined expression manifold, causing uncanny artifacts. For example, the presence of hand in HMC images leads to out-of-distribution (OOD) generalization issues by the facial encoder, which produces OOD latent codes that the UPM model fails to generalize to. This suggests the necessity of future research on the OOD generalization for HMC-based avatar animation. Finally, our system currently does not handle tongue tracking well, primarily due to the confusion between tongue, lips, and skin under the extreme lighting, which leads to error in the HMC-avatar correspondence generation stage (i.e., the labels themselves are imperfect). We believe it can be improved by better segmentation annotation.

\subsection{Ablations}
\label{subsec:experiments-ablations}

In this subsection, we dissect the distinct algorithmic designs, contributions and data choices we made via multiple ablative experiments to gauge their influence. Specifically, we scrutinize the ensuing aspects: training objectives, the choice of the feature-level calibration module $\phi_\text{cal}$, self-supervision strategies, and the identity scaling performance.

\begin{table}
\ra{1.3}
\caption{\textbf{Ablation on training loss for the universal facial encoder model $\mathcal{E}_\text{univ}$.} No SSL pretrained weights are used.}
\resizebox{0.48\textwidth}{!}{
\begin{tabular}{l|ccc}
\toprule
& Photometric $L_1$ $\downarrow$ & Mesh $L_2$ $\downarrow$ & Mouth Shape $L_2$ $\downarrow$ \\
\midrule
Code loss & 2.482 & 1.074 & 2.415 \\
Perceptual loss & \cellcolor{green!90!black}\textbf{2.413} & \cellcolor{green!90!black}\textbf{1.071} & \cellcolor{green!90!black}\textbf{2.355} \\
\bottomrule
\end{tabular}}
\label{tab:experiments-ablation-training-loss}
\end{table}

\begin{table}
\ra{1.3}
\caption{\textbf{Ablation on the choice of expression feature calibration $\phi_\text{cal}$.} No SSL pretrained weights are used.}
\resizebox{0.48\textwidth}{!}{
\begin{tabular}{l|ccc}
\toprule
& Photometric $L_1$ $\downarrow$ & Mesh $L_2$ $\downarrow$ & Mouth Shape $L_2$ $\downarrow$ \\
\midrule
Cross-attention & \cellcolor{green!10}2.501 & \cellcolor{green!10}1.157 & \cellcolor{green!10}2.523 \\
Self-attention & 2.736 & 1.200 & 2.908 \\
Pooling Aggregation & \cellcolor{green!30}2.469 & \cellcolor{green!30}1.121 & \cellcolor{green!30}2.475 \\
Input Early Fusion & 2.610 & 1.162 & 2.687 \\
Late Fusion & \cellcolor{green!90!black}\textbf{2.413} & \cellcolor{green!90!black}\textbf{1.071} & \cellcolor{green!90!black}\textbf{2.355} \\
\bottomrule
\end{tabular}}
\label{tab:experiments-ablation-calibration-architecture}
\end{table}

\begin{table}
\ra{1.3}
\caption{\textbf{Ablation on different self-supervised learning approaches.} We compare different pre-training strategies by comparing their performance on the downstream finetuning using high-quality labels from photorealistic avatars.}
\resizebox{0.48\textwidth}{!}{
\begin{tabular}{l|ccc}
\toprule
& Photometric $L_1$ $\downarrow$ & Mesh $L_2$ $\downarrow$ & Mouth Shape $L_2$ $\downarrow$ \\
\midrule
No SSL & 2.413 & 1.071 & \cellcolor{green!10}2.355 \\
SimCLR (contrastive) & 2.436 & 1.090 & 2.442 \\
ImageNet Pre-training & \cellcolor{green!10}2.408 & \cellcolor{green!10}1.067 & 2.379 \\
MAE Orig. View Recon & \cellcolor{green!30}2.385 & \cellcolor{green!30}1.055 & \cellcolor{green!30}2.234 \\
MAE Novel View Recon. & \cellcolor{green!90!black}\textbf{2.309} & \cellcolor{green!90!black}\textbf{1.023} & \cellcolor{green!90!black}\textbf{2.083}\\
\bottomrule
\end{tabular}}
\label{tab:experiments-ablation-ssl-strategy}
\end{table}

\paragraph{Training objectives.} We assess the importance of the perceptual losses ($\mathcal{L}_\text{photo}$, $\mathcal{L}_\text{geo}$) we use, in comparison to using direct latent code loss $\|\hat{\mathbf{e}} - \mathbf{e}^\star\|_2$ (which resembles the blendshape coefficient loss that prior works have used for animating blendshape face models). The results are shown in Table~\ref{tab:experiments-ablation-training-loss}. Using perceptual training losses leads to overall better quantitative performances, but we do note that even without them, the results are highly competitive, already surpassing many current SOTA approaches that \emph{\textbf{are}} supervised with perceptual losses (see Table~\ref{tab:experiments-full-baseline-comparison}). This demonstrates the improvement brought by our architectural design (e.g., multi-expression calibration mechanism). 

\paragraph{Choice of feature-level calibration module.} As we discussed in \S\ref{subsec:multi-expression}, the feature-level calibration module $\phi_\text{cal}$ mixes current expression feature with anchor expression features, and could take different architectural forms. We compare the following options in Table~\ref{tab:experiments-ablation-calibration-architecture}: 1) self-attention (where current \& anchor expression feature maps are concatenated and fed into a Transformer encoder)~\citep{dosovitskiy2020image}; 2) cross-attention (anchor features serve as keys and values, and current features as queries~\citep{vaswani2017attention}); 3) pooling-based aggregation (pool the anchor features and concatenate with current frame)~\citep{qi2017pointnet}, 4) early fusion (raw images concatenation); and 5) late-fusion (feature-level concatenation followed by shallow convolution blocks taken from MobileNetv3~\citep{howard2019searching}). Compared to attention- or pooling-based calibration architectures, we found the late-fusion-based calibration module to be the most effective (with overall best photometric loss) and most efficient (as attention layer's computation cost scales quadratically with the feature map resolution and number of anchor expressions). 

\paragraph{Self-supervision strategies.} While we adopted a masked-autoencoder (MAE)~\citep{he2022masked} based novel-view reconstruction as a pretext task for SSL on unlabeled data, we compare with other pre-training strategies that have shown to be beneficial for computer vision tasks. Specifically, we compare our approach with a similar setting where we use the original HMC views on the tracking headset (i.e., not novel views), an ImageNet pre-trained model, and a SimCLR-based contrastive learning with InfoNCE loss~\citep{chen2020simple} (with large batch size (e.g., 512) where we consider both different views (e.g., same expression frame but from a different camera) and different augmentations of the same image as positive examples). The results of these different SSL pre-trained weights on downstream supervised fine-tuning are shown in Table~\ref{tab:experiments-ablation-ssl-strategy}. Neither ImageNet pretraining nor contrastive learning lead to obvious improvements over the baseline. For ImageNet pre-trained weights, we suspect the domain change is too large (i.e., from ImageNet-1K~\citep{deng2009imagenet} to HMC data) for the transfer learning. For contrastive learning, we hypothesize this is a result of the way positive/negative examples are currently defined. While we follow the standard practice of using the multi-view representation of the same frame as positive examples~\citep{chen2020simple,zbontar2021barlow}, this does \emph{not} prevent the model from overfitting on identity information unrelated to expression. For instance, a model could learn to compare facial attributes like face shape, rather than expressions, and yet still achieving low contrastive loss. Ideally, we should require cases like ``person A jaw open max'' and ``person B jaw open max'' to also be positive HMC input pairs. However, this could require further careful annotation on expression alignment, which could be challenging (see our discussion in \S\ref{sec:discussion}), and we leave it to future work. Overall, our proposed novel view reconstruction method leads to the largest improvement out of all settings here.

\begin{figure}
  \centering
  \includegraphics[width=0.47\textwidth]{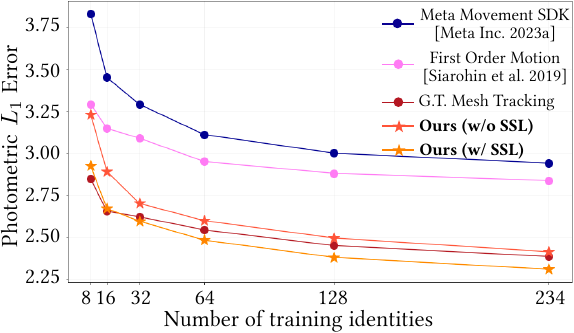}
  \caption{\textbf{Identity scaling.} We study how the performance of different algorithms change as a function of \emph{number of identities with high-quality labels} that are used for training the universal facial encoder. We pick 8, 16, 32, 64, 128, and 234 (i.e., all) training subjects, while keeping the 32-subject test set intact in all the settings.}
  \label{fig:experiments-scaling-law}
\end{figure}

\paragraph{Identity scaling.} We analyze in Fig.~\ref{fig:experiments-scaling-law} how these algorithmic components and input representations influence the generalization performance (i.e., on unseen subjects) as a function of the number of labeled training identities used. We specifically also add the hypothetical setting where groundtruth mesh is used as an input (see Table~\ref{tab:experiments-full-baseline-comparison}). An interesting phenomenon is that in the low-data regime, both keypoint warp-field-based approach (\textcolor{pink}{pink} line) and mesh-tracking based approach (\textcolor{brown}{brown} line; again, assuming perfect mesh tracking) obtain quite competitive performance. However, our universal encoding model (\textcolor{red}{red} line) with multi-expression calibration catches up on generalization performance very quickly, and with the help of self-supervised pre-training (\textcolor{orange}{orange} line), even surpasses the performance of the contrived mesh-tracking method (while performing almost equally well in the low-data regime).

\section{Broader Impact}
\label{sec:broader-impact}

The ability to accurately animate digital representations of real people simultaneously offers the possibility to enhance remote communication while presenting an opportunity for abuse and misuse. Key areas of risk that stand out include the risks of impersonation, deepfakes, and appearance tampering. Of these, the potential loss of trust in the communication medium from online impersonation is arguably the most important. It is also that to which our work is most relevant. 

Our method's ability to generalize to new users using only a small number of anchor expressions presents an elevated risk of abuse. In comparison, avatar creation typically requires significantly more data of the target individual to generate a convincing \emph{digital double}~\cite{cao2022authentic}. An adversary with physical access to a user’s device could drive the victim’s avatar. 
To mitigate this threat, strong authentication and verification of identity is required to control access at runtime. Existing technologies such as a PIN created during enrollment or biometric authentication modalities can be implemented. Similarly, the expression-calibration images used in our method could also be collected only once during enrollment and reused for all future sessions.

Although traditional approaches to authentication would mitigate many attack vectors, they do not prevent collusion by the owner of the avatar, who may deliberately allow another individual to take their place during HMC-portions of enrollment (e.g., to cheat on a proctored exam or interview for a job). While this challenge is present with most avatar representations, authentic photorealistic representations, such as codec avatars, afford a distinct solution to prevent misuse; biometric matching between the individual wearing the headset and the avatar they are driving. Stylized representations and significant appearance augmentations, while imparting an enhanced level of expressive control, do not provide appearance authenticity and prevent the avatar's use as a means for authentication. Our ground truth generation technique (Sec.~\ref{subsec:rosetta}) relies on the ability to match decoders to HMC images, and extensions of these techniques may be applicable to live verification approaches as well. 

Finally, in cases where completely securing the system against inauthentic usage can not be achieved, platform-level mitigation strategies employed in other domains, such as banking or social media, will be needed. Functionality such as user-notification of avatar use and enabling users to report suspected impersonation will be important detection tools.

\section{Discussion and Conclusions}
\label{sec:discussion}
In this work, we present an end-to-end universal facial encoding system for photorealistic codec avatars that is able to work directly from an off-the-shelf VR headset with challenging camera views, donning variation, and illumination conditions. Specifically, we first introduce an enhanced parameterization that is able to address limitations introduced by extreme illumination changes by explicitly modeling light codes. This is able to provide more precise HMC-avatar correspondences.

Moreover, we introduce a lightweight multi-expression calibration architecture that leads to substantial gains in interpolating accurate facial motions. With the help of multiple anchor expressions, we show that an end-to-end facial encoding system (which directly takes in raw images) improves significantly in capturing individual idiosyncrasies and resolving the muted expression problem (due to the oblique camera view). Importantly, this comes at a nominal cost to inference-time efficiency.

We also present a new self-supervised learning (SSL) pipeline for learning facial encoding systems from large-scale \emph{unlabeled} head-mounted camera data, by leveraging novel-view reconstruction with MAEs. This is in contrast to prior works which have relied on the need to build 3D face model and high-quality labels (e.g., blendshape coefficients or latent representations). Our results show that a subsequently finetuned facial encoding model can animate photorealistic avatars in realtime (e.g., in a two-way call) with high fidelity, and achieves the best quantitative results on several metrics across the board (e.g., by over 20\% improvements over existing approaches). Qualitatively, our approach can capture extreme and subtle expression changes well (i.e., sensitive to nuanced signals from the input images), while being simultaneously robust to certain out-of-distribution input changes (e.g., wearing glasses).

Yet the system still faces multiple challenges. First, the exact concept of \emph{expression alignment} is a somewhat ill-posed problem for photorealistic facial motions. For example, each person can have multiple states of ``neutral'' expressions, all with subtle differences. Then it is unclear how these multiple ``neutral'' states of a person consistently retarget to the ``neutral'' those of a different person (e.g., should they be more like the source identity, or the target identity?). This creates challenge as we encode to a \emph{universal} latent space that can be used for re-targeting. Second, we believe it would be interesting to add new modalities to the model architecture to address some of the sensing challenges. For example, with synchronized audio from the microphone, the universal facial encoding system may be able to improve on the lip motion during speech.

To summarize, we have presented the first generalizable, accurate, and real-time universal facial encoding solution that works for off-the-shelf consumer VR headsets, which present multiple sensing challenges. Our encoding system is able to leverage large-scale self-supervised pre-training, and produces results significantly outperforming any prior approaches on photorealistic avatars.

\begin{acks}
We are immensely grateful to Zhengyu Yang for the instant codec avatar models; to Jiajun Tan for the live demo system setup; to Julieta Martinez Covarrubias for the dataset release support; to Autumn Trimble, Julia Buffalini, Kevyn McPhail and Yu Han for the video recordings; to Shoou-I Yu for the avatar asset processing. 
\end{acks}

\bibliographystyle{ACM-Reference-Format}
\bibliography{main}

\newpage
\appendix

\section*{Appendix}
\section{Search for the Anchor Expressions}
\label{appendix:anchor-search}

Our search for the best anchor expressions comes at two criteria: 1) they are common and unambiguous in interpretation; and 2) an encoder model with these anchor expressions achieve the lowest photometric error. Therefore, from the HMC capture segments (see \S\ref{sec:experiments}), we first select a candidate set $E$ of 15 expressions (e.g., ``neutral'', ``maximal jaw dropping'', ``wide eyes'', ``closed eyes'', etc.) that satisfy criteria (1). Then, we perform a heuristic (greedy) search that is based on the beam search algorithm~\citep{steinbiss1994improvements}. The pseudocode of the algorithm is presented in Alg.~\ref{alg:app-beam-search}.  

\begin{algorithm}
\caption{Greedy Search for Calibration Expressions}\label{alg:app-beam-search}
\begin{algorithmic}
\Require Encoder model $M(\cdot, \mathbf{y})$, beam $\mathbf{B}$, beam width $k$, visited set $S$, candidate expressions set $E$, test-set evaluation $\mathsf{Test}(\cdot)$.
\State $\mathbf{B} \gets \{\}$
\State Res $\gets$ []  \Comment{Cache for all visited nodes and test results}
\While{($\mathbf{B} = \varnothing$) or ($\exists b \in \mathbf{B} \text{ s.t. } |b| \geq 6$)}
\For{$b \in \mathbf{B}$, $e \in E$}
    \If{$b \cup \{e\} \notin S$}  
        \State 1. $M' \gets M(\cdot, b \cup \{e\})$, train $M'$  \Comment{New candidate}
        \State 2. Res $\gets$ Res + \big[$(\mathsf{Test}(M'), b \cup \{e\})$\big]
        \State $S \gets S \cup \{b \cup \{e\}\}$
    \EndIf
\EndFor
\State Res' $\gets$ sort(Res)[:$k$] \Comment{Pick $k$ lowest error settings}
\State $\mathbf{B}' \gets \{\text{res[1] for res} \in \text{Res'}\}$  \Comment{Tentative new beam}
\If{$\mathbf{B}' = \mathbf{B}$}
    \State \textbf{STOP} \Comment{No improvement from adding more expressions}
\EndIf
\State $\mathbf{B} \gets \mathbf{B}'$
\EndWhile

\State \textbf{return} $\mathbf{B}$
\end{algorithmic}
\end{algorithm}

\begin{table}
\centering
\setlength{\tabcolsep}{3.5pt}
\ra{1.2} 
\caption{\textbf{Additional dataset and hyperparameter details}. Unless stated specifically for SSL, the hyperparameters are for the supervised universal facial encoder training.}
\resizebox{0.48\textwidth}{!}{\begin{tabular}{c|c|c}
\toprule
\multirow{6}{*}{Dataset} & Num. of frames (train) & 70M \\
& Num. of frames (test) & 10M \\
& Num. of frames (train, SSL) & 2.9B \\
& HMC input resolution & 192 $\times$ 192 \\
& HMC input augmentations & CutMix, rotation, scale \\
& VR Headset FPS & 72 \\
\cmidrule{1-3}
\multirow{13}{*}{Hyperparameters} & Architecture for $\phi_\text{exp}$ & MobileNetV3 \\
& Learning rate schedule & Cosine annealing \\
& Number of training steps & 150000 \\
& Number of training steps (SSL) & 400000 \\
& Batch size & 16 \\
& Parameter optimizer & AdamW \\
& Weight decay & $10^{-5}$ \\
& Weight decay (SSL) & $5 \cdot 10^{-6}$ \\
& MAE mask ratio (SSL) & 0.9 \\
& Identity embedding $\mathbf{u}^\text{id}$ shape (SSL) & $\mathbb{R}^{6 \times 6 \times 192}$ \\
& Random rotation degree ($\pm$) & 15 \\
& Number of CutMix patches & 3 \\
& Smoothness $\sigma$ (SSL, see Eq.~\ref{eq:ssl-canonicalization}) & 40 \\
\bottomrule
\end{tabular}}
\label{tab:appendix-hyperparameters}
\end{table}

\begin{table*}
\centering
\setlength{\tabcolsep}{3.5pt}
\ra{1.28} 
\caption{\textbf{Symbols used in this paper for different models and features}.}
\resizebox{\textwidth}{!}{
\begin{tabular}{c|c|c|p{0.95\linewidth}}
\toprule
Categories & Symbols & Dimensions & Meanings \\
\cmidrule{1-4}
\multirow{21}{*}{Model} & $\mathcal{E}_\text{exp}$ & - & \emph{Expression encoder} that generates expression codes to train the universal prior model (UPM; i.e., the decoder), using view-averaged texture $\textbf{T}_\text{exp}$ and geometry images $\textbf{G}_\text{exp}$. \\
 & $\mathcal{E}_\text{id}$ & - & \emph{Identity encoder} that generates per-subject bias maps $\Theta_\text{id}$ from a neutral texture map $\textbf{T}_\text{neu}$ and a neutral geometry image $\textbf{G}_\text{neu}$. \\
 & $\mathcal{D}$ & - & \emph{Universal Prior Model (UPM)}; a.k.a. the decoder model. It takes disentangled latent codes (containing expression information) and the person-specific bias maps $\Theta_\text{id}$ to produce volumetric slabs $\textbf{M}$ and geometry $\textbf{G}$ of the 3D avatar face. \\
 & $R_\text{vol}$ & - & \emph{Volume ray-marching module} which evaluates rays of light passing through the volumetric slabs $\textbf{M}$ produced by the decoder $\mathcal{D}$. It produces a volumetric rendering image $I_\text{vol}$ given a viewpoint $\mathbf{v}$. \\
 & $R_\text{mesh}$ & - & \emph{Mesh renderer with rasterization} that renders the avatar from a specified viewpoint $\mathbf{v}$ using decoder output $\textbf{G}$ and $\textbf{T}$. This is used in establishing HMC-avatar correspondences only, where we need 2D style transfer (see \S~\ref{subsec:rosetta}). \\
 & $\mathcal{F}$ & - & \emph{Style transfer module} which converts the RGB decoder texture to monochrome (but without lighting information). \\
 & $\mathcal{E}_\text{cor}$ & - & \emph{Personalized HMC attribute encoder} which produces expression $\mathbf{e}$, HMC camera viewpoint $\mathbf{v}$, gaze information $\mathbf{g}$ and lighting code $\mathbf{l}$ which are optimized to become the HMC-avatar correspondence. \\
 & $\mathcal{E}_\text{univ}$ & - & \emph{Universal facial encoder} which takes the HMC camera image of any subject (along with anchor expressions, if any), and produce the expression latent codes $[\hat{\mathbf{e}}, \hat{\mathbf{g}}]$. \\
 & $\mathcal{E}_\text{univ}'$ & - & \emph{Pre-training universal facial encoder} used in SSL for novel-view reconstruction. Finetuned to become $\mathcal{E}_\text{univ}$. \\
 & $\phi_\text{exp}^\text{\{U,L\}}$ & - & \emph{Face-region-specific expression feature extractor} that extracts low-level (and potentially without calibrated) image features. U/L stand for upper/lower-face regions, respectively. \\
 & $\phi_\text{cal}^\text{\{U,L\}}$ & - & \emph{Feature-level calibration modules} which calibrate the expression feature map of the current frame with respect to $M$ anchor expressions. \\
 & $\mathcal{U}^\text{\{up,down\}}$ & - & \emph{The up/down-sampling portion of U-Nets} that performs novel-view reconstruction given a highly masked output, expression feature $\mathcal{E}_\text{univ}(\mathbf{x})'$, and an identity embedding $\mathbf{u}^\text{id}$. \\
\cmidrule{1-4}
\multirow{10}{*}{Feature} & $\Theta_\text{id}$ & - & \emph{Per-subject} bias maps that are injected to different UPM decoder blocks during the decoding process. \\
& $\textbf{T}_\text{neu,exp}$ & - & \emph{View-averaged texture} of current or neutral expression frame(s), respectively, produced from face images of a multi-camera system.  \\
& $\textbf{G}_\text{neu,exp}$ & - & \emph{Geometry image} registered from face images of a multi-camera dome system. \\
& $\textbf{M}$ & - & \emph{Volumetric slabs}~\citep{lombardi2021mixture} produced by the UPM decoder. \\
& $\mathbf{e}$ & $\mathbb{R}^\text{256}$ & \emph{Expression latent code} in a universal latent space, produced by encoders and then decoded by UPM to form a 3D avatar model. \\
& $\mathbf{g}$ & $\mathbb{R}^6$ & \emph{Gaze vectors} for the explicit eye model (EEM)~\citep{schwartz2020eyes}, 3-dimensional for each eye (so 6-dim in total). \\
& $\mathbf{l}$ & $\mathbb{R}^4$ & \emph{Lighting code} produced during the HMC-avatar correspondence establishment. \\
& $\mathbf{u}^\text{id}$ & $\mathbb{R}^p$ & \emph{Identity embedding} used to inform the U-Net in SSL on identity-specific information during reconstruction (See \S~\ref{subsec:ssl}). \\
& $\mathbf{v}$ & $\mathbb{R}^6$ & \emph{Viewpoint} for rendering the avatar faces (using volumetric or mesh-based renderer). \\
\bottomrule
\end{tabular}}
\label{tab:appendix-symbols}
\end{table*}

Intuitively, we keep a beam $\mathbf{B}$ of maximum size $k$ (called the beam width; empirically we use $k=3$) that contains the most promising $k$ nodes (i.e., expression combinations). Then, at each level of the search tree, we generate all successors $b \cup \{e\}$ of the expression combination states (i.e., new possible combinations), train the encoder model $\mathcal{E}_\text{univ}=M$ with these successor nodes, and sort them by the test-set evaluation results. After that, importantly, we only store and proceed to the next level of search with the $k$ best nodes. We repeat the process, until we reach a promising node with 6 expressions, or if the beam stop updating (meaning the improvement brought by adding anchor expressions saturates).

Empirically, we find it almost always useful (for error mitigation) to add anchor expressions, though at a diminishing return. Thus, we have not observed in practice when Alg.~\ref{alg:app-beam-search} early stops.

\section{Additional Statistics and Hyperparameters}
\label{appendix:additional-stats}

In Table~\ref{tab:appendix-hyperparameters}, we provide dataset statsitics and hyperparameters used in model training, in addition to the ones outlined in \S~\ref{sec:experiments}.

\section{Table of Symbols}
\label{appendix:table-of-symbols}

In Table~\ref{tab:appendix-symbols}, we provide a comprehensive table for the different symbols we used in \S~\ref{sec:decoder} and \S~\ref{sec:encoder} for better notation clarity.

\end{document}